\documentclass{article}

\usepackage{silence}
\usepackage{hyphenat}
\usepackage{style/macros}
\usepackage{style/opendir_macro}

\usepackage{natbib}
\usepackage{subcaption}
\usepackage{caption}

\usepackage{libertine}
\usepackage[margin=1in]{geometry}

\usepackage{titling}
\newcommand{\titlefontsize}{20pt}    % title font size
\pretitle{\par\centering\fontsize{\titlefontsize}{1.2\titlefontsize}\selectfont}
\posttitle{\par}

\title{There Will Be a Scientific Theory of Deep Learning}
\newcommand{\authorpt}{10.5}
\newcommand{\authorlead}{12.2}

\newcommand{\authorgap}{2em}
\newcommand{\authortopgap}{2em}
\newcommand{\authorshift}{5mm}        % move authors DOWN (0mm to disable)

\newcommand{\authortopwidth}{0.30\linewidth}
\newcommand{\authortopcolgap}{0.6em}

\newcommand{\authorcolwidth}{0.26\linewidth}
\newcommand{\authorcolgap}{3em}

\newcommand{\authornameaffilvspace}{-0.05em}

\newcommand{\authorxshift}{-4mm}      % nudge whole author block (negative = left)
\newcommand{\authorfont}{\fontsize{\authorpt}{\authorlead}\selectfont}

\newcommand{\TopAuthor}[2]{%
  \parbox[t]{\authortopwidth}{\centering
    {\bfseries #1}\\[\authornameaffilvspace]
    {\itshape #2}%
  }%
}
\newcommand{\AuthorCell}[2]{%
  \parbox[t]{\authorcolwidth}{\centering
    {\bfseries #1}\\[\authornameaffilvspace]
    {\itshape #2}%
  }%
}

\newcommand{\CenteredLine}[1]{\makebox[\linewidth][c]{\hspace*{\authorxshift}#1}}

\author{%
\authorfont%
\begin{minipage}{\linewidth}\centering
\vspace*{\authorshift}%
\CenteredLine{%
\begin{tabular}{@{}c@{\hspace{\authortopcolgap}}c@{}}
\TopAuthor{Jamie Simon$^*$}{UC Berkeley and Imbue} &
\TopAuthor{Daniel Kunin}{UC Berkeley}
\end{tabular}}\\[\authortopgap]
\CenteredLine{%
\begin{tabular}{@{}c@{\hspace{\authorcolgap}}c@{\hspace{\authorcolgap}}c@{}}
\AuthorCell{Alexander Atanasov}{Harvard University} &
\AuthorCell{Enric Boix-Adser\`a}{University of Pennsylvania} &
\AuthorCell{Blake Bordelon}{Harvard University}\\[\authorgap]
\AuthorCell{Jeremy Cohen}{Flatiron Institute} &
\AuthorCell{Nikhil Ghosh}{Flatiron Institute} &
\AuthorCell{Florentin Guth}{NYU and Flatiron Institute}\\[\authorgap]
\AuthorCell{Arthur Jacot}{New York University} &
\AuthorCell{Mason Kamb}{Stanford University} &
\AuthorCell{Dhruva Karkada}{UC Berkeley}\\[\authorgap]
\AuthorCell{Eric J. Michaud}{Astera Institute} &
\AuthorCell{Berkan Ottlik}{University of Pennsylvania} &
\AuthorCell{Joseph Turnbull}{UC Berkeley}\\
\end{tabular}}%
\end{minipage}%
}

\date{}
\begin{document}
\maketitle
\vspace{-3mm}
\begin{abstract}

In this paper, we make the case that a scientific theory of deep learning is emerging.
By this we mean a theory which characterizes important properties and statistics of the training process, hidden representations, final weights, and performance of neural networks.
We pull together major strands of ongoing research in deep learning theory and identify five growing bodies of work that point toward such a theory:
\begin{enumerate}
    \item solvable idealized settings that provide intuition for learning dynamics in realistic systems;
    \item  tractable limits that reveal insights into fundamental learning phenomena;
    \item simple mathematical laws that capture important macroscopic observables;
    \item theories of hyperparameters that disentangle them from the rest of the training process, leaving simpler systems behind; and
    \item universal behaviors shared across systems and settings that clarify which phenomena call for explanation.
\end{enumerate}

Taken together, these bodies of work share certain broad traits: they are concerned with the dynamics of the training process; they primarily seek to describe coarse aggregate statistics; and they emphasize falsifiable quantitative predictions.
We argue that the emerging theory is best thought of as a \textit{mechanics} of the learning process, and suggest the name \textit{learning mechanics}.
We assert that learning mechanics should be a mathematical theory, grounded in first-principles calculations that closely predict empirics, reliant on well-tested approximations and assumptions, aiming for broad impact across the machine learning stack once it reaches maturity.

We discuss the relationship between this mechanics perspective and other approaches for building a theory of deep learning, including the statistical and information-theoretic perspectives.
In particular, we anticipate a symbiotic and mutually supportive relationship between learning mechanics and the developing discipline of mechanistic interpretability.
Where mechanistic interpretability aims to be the biology of deep learning, learning mechanics should aspire to be its physics, mirroring the complementary relationship between biology and physics in the natural sciences.

We also review and address common arguments that fundamental theory will not be possible or is not important.
We conclude with a portrait of important open directions in learning mechanics and advice for beginners.
We host further introductory materials, perspectives, and open questions at \websiteurl.

\end{abstract}

\numberlessfootnote{$^*$Correspondence to \texttt{james.simon@berkeley.edu}.}
% and a little old man in a hut atop Mount Shasta reachable only by two days' hike when the snow is favorable, but if you do he'll let you in and give you tea and answer your questions.}

\setcounter{footnote}{0}
\newpage

% SECTION 1
\section{Introduction}
\label{sec:intro}

% \jamie{tone goals for the intro + rest of paper, relevant again for the rewrite:
% \begin{itemize}
%     \item be clear. be succinct.
%     \item be other-oriented: ask what we can do for others.
%     \item come off as likable. don't sound pompous or full of ourselves. don't be afraid to be controversial, but don't offend.
%     \item be inspiring. write like we weren't going to be able to do the research ourselves and had to inspire other people to do it.
%     \item convey optimism and hope.
% \end{itemize}
% }

Deep learning is famously a black-box learning method, the most powerful, most inscrutable, and now most technologically important member of the machine learning pantheon.
Properly trained, neural networks learn to perform a wide array of tasks with superhuman performance, but we have no unified scientific framework that explains why or how.
Motivated by both scientific curiosity and the promise of practical engineering benefit, the effort to put rigorous mathematical and scientific backing behind this applied discipline has spanned decades.
Despite some progress, however, our understanding remains primitive: neural networks are still trained using methods discovered largely through trial and error rather than first principles, and theory plays little role in the day-to-day practice of deep learning.
The challenge has only compounded as practice has advanced, and in the era of large language models and diffusion models, the mysteries are arguably deeper than they were one or two decades ago.
Will we ever understand?

\vspace{2mm}
\begin{tcolorbox}
[sharp corners, colback=white, colframe=black]
    \textbf{
    \hspace{-1.5mm}
    This paper makes the case that, yes, there will be a scientific theory of deep learning; that we can see pieces of this theory starting to emerge; and that this theory will take the form of a \textit{mechanics} of the learning process.
    }
\end{tcolorbox}
\vspace{2mm}

The questions driving deep learning theory have changed over time, and to understand where the field is going, it is useful to first look back at how we got here.
Deep learning theory is as old as machine learning itself, with roots in the McCulloch--Pitts neuron and the perceptron in the middle of the last century.
The earliest theoretical questions in machine learning were about expressivity: what functions can simple models represent, and how can they be learned from data?
As learning came to be understood as a statistical problem, and simple learning systems found practical success, the theoretical focus shifted to ask: when does learning from finite samples generalize?
This gave rise to \textit{classical learning theory}, including statistical and computational/PAC learning theory.
Paired with \textit{classical optimization theory,} these frameworks gave clean end-to-end guarantees of the optimization and generalization of simple learning systems.
In parallel, a classical tradition of the \textit{statistical physics of machine learning} developed satisfying theories of the average-case behavior of simple models.

While these classical theories built a strong foundation for understanding learning, the rise of deep learning through multilayer networks, backpropagation, and increasing scale in both data and compute exposed limitations in their explanatory power.
Neural networks are complex, nonconvex, and overparameterized (in contrast with the simple, convex, parsimonious models for which classical learning theory excels), and they optimize and generalize better than these classical approaches can guarantee or explain.
Furthermore, it became clear that neural networks were not merely fitting data or achieving low training error, they were learning structured internal representations and displaying striking regularities across tasks and scales.
The classical questions of performance and efficiency remained important, but answering them would first require understanding a new host of phenomena shaped both by the dynamics of neural networks through training and the structure of the data they are trained on.

This marked a transition in which deep learning theory changed in character from a largely \textit{mathematical} study of what is possible to a truly \textit{scientific} effort to describe, explain, and ultimately predict the behavior of complex empirical systems.
New scientific endeavors often start with an empirical tension in which nature presents something interesting we cannot predict or explain with existing tools, and although neural networks are artificial computational systems, this same scientific tension is present here.
We should thus approach this task as scientists, embracing empirics, seeking unifying principles, and identifying recurring motifs.
We should also expect the path forward to look more like the development of a scientific field than the development of a mathematical one.

The purpose of this paper is to convince the reader that this scientific tension is gradually giving way to a scientific theory which resolves it.
In \cref{sec:reasons_why}, we pull together major strands of ongoing research and identify five lines of evidence that such a theory is emerging:

\begin{enumerate}
    \item there are a growing number of \textit{analytically solvable settings} in which learning is fully captured by simple mathematics, including but not limited to deep linear networks and kernel methods (\cref{sec:reason-dynamics});
    \item there exist \textit{useful limits}, including limits of infinite width and depth, that provide insight into fundamental learning behaviors (\cref{sec:reason-limits});
    \item in many cases, \textit{simple empirical laws} suffice to capture meaningful macroscopic statistics, including test-time performance and loss-landscape sharpness (\cref{sec:reason-measurable});
    \item many of the hyperparameters governing optimization can be \textit{disentangled and understood}, leaving behind a simpler effective dynamical system (\cref{sec:reason-hps}); and
    \item as applied deep learning has scaled up and converged on best practices, \textit{universal phenomena} have increasingly appeared across settings and tasks (\cref{sec:reason-universal-behavior}).
\end{enumerate}

% What will this emerging science look like when fully developed?
% 
These lines of research broadly share several overarching characteristics: they are concerned with the \textit{dynamics of the training process}; they primarily seek to describe \textit{coarse aggregate statistics} of learning; and they emphasize \textit{accurate average-case predictions} over rigorous worst-case bounds.
In this sense, the emerging scientific theory of neural networks appears to have much in common with theories in physics such as classical mechanics, continuum mechanics, statistical mechanics, and quantum mechanics.
We argue that this emerging theory is best understood as a \textit{mechanics of learning}.

\subsection{What's in a mechanics?}

Mechanics is the branch of physics studying how forces acting on objects determine their movement through space and time.
Neural network learning can be thought of in this way: much as an object moves continuously through physical space, learning involves a model moving through parameter space via discrete updates.
In the physical sciences, forces come from interactions between components of a system.
Similarly, the process of deep learning is shaped by interactions between the parameters, dataset, task, and learning rule.
In physics, these forces are mediated by fields; in deep learning, they are mediated by gradients.
In physics, systems settle into equilibria at local minima of a potential determined by internal interactions and external constraints; analogously, neural networks converge to local minima of a loss landscape shaped by their architecture and training data.
While the systems under study are very different, since the key problems of both are essentially about movement and interaction, we might expect some features of the resulting sciences to be shared.

% \dhruva{I think the table shown later should actually referenced here and not there. We probably don't need to spell out the whole 5-part analogy in the main text; maybe some of these details can be folded into the table caption. I think it's fine in the main text to say ``not just speculation, deep similarities are shown in the table, by analogy we call our intended theory LM."}
These analogies are not just speculation: we can see these similarities reflected in the lines of research listed above.
All branches of mechanics (and especially classical mechanics) develop a library of analytically solvable settings to gain intuition; so too does learning mechanics.
All branches of mechanics use limits as simplifying tools; so too does learning mechanics.
Continuum and statistical mechanics, the branches which most directly deal with large numbers of interacting components, describe zoomed-out summary statistics rather than the motion of every particle; this has also proven a useful approach in dealing with the complexity of deep learning.
Every physical system has one or more system parameters (characteristic scales, coupling constants, etc.) affecting its behavior, and some techniques for treating these are essentially the same as those used to study hyperparameters in deep learning.
Finally, physics is full of cases in which the same phenomena show up in very different settings, and similarly we see universal behavior emerging across deep learning systems.

All considered, the emerging science shares deep similarities with established branches of mechanics.
By analogy to classical, continuum, statistical, and quantum mechanics, we suggest the intended theory be called \textit{learning mechanics}.

\paragraph{Seven desiderata for learning mechanics.}

% Let us assume for a moment that we want a mechanics of learning and consider the challenge of developing it.
% %
% This is a big undertaking, and so we should be clear about what it is we want from this research program.

We should be clear at the outset what we want from a mechanics of learning.
Assessing how mature branches of mechanics were motivated, developed, and succeeded, we can see what sort of goals to aim for.
Here are seven desiderata for this research program:

\begin{enumerate}[itemsep=1ex]
    \item
    Learning mechanics should be \textit{fundamental}, proceeding logically from a first-principles description of neural network training.
    Interim assumptions about network weights, dynamics, and performance will be useful tools, but they should ultimately be explained from first principles.
    \item
    Learning mechanics should be \textit{mathematical}, making unambiguous quantitative statements about important properties of neural networks.
    No mechanics is a qualitative science; neither will be learning mechanics.
    \item
    Learning mechanics should be \textit{predictive}, making claims supported by simple, repeatable empirical measurements.
    We have excellent experimental control of our system, and every major development should be unambiguously verified in experiment.
    \item Learning mechanics should be \textit{comprehensive}, describing aspects of neural networks' training process, hidden representations, and final weights in a single picture.
    It is worth emphasizing that this theory will not --- and should not --- aim to describe everything.
    A map at the full resolution of the world would be the size of the world and thus of little use.
    What we seek instead is a theory that operates at the right level of resolution --- one that sacrifices detail in favor of insight.
    \item Learning mechanics should be \textit{intuitive}, being simple, illuminating, and satisfying in its demystification of deep learning.
    Like physics, learning mechanics should strive for simple insight over technical complexity.
    \item Learning mechanics should be \emph{useful}, serving as the scientific foundation for applied deep learning as physics does for other forms of engineering.
    Concrete goals should include greatly reducing the need for hyperparameter tuning, giving predictive tools for dataset design, and providing rigorous foundations for AI safety work.
    \item Finally, learning mechanics should be \textit{humble}, being solid in what it describes and explicit about what it cannot.
    Every branch of physical science has a regime of applicability outside of which it breaks down, and these boundaries are taught together with the science so that it may be used reliably.
    We anticipate the mechanics of learning applicable to realistic deep learning will break down in many small-scale, handcrafted, or otherwise special cases, and this is the price we will pay for the right simple picture in the regimes we care about.
\end{enumerate}

A mechanics of learning with these virtues --- one that is fundamental, mathematical, predictive, comprehensive, intuitive, useful, and humble --- would be transformative, paradigm-setting.
We expect such a theory would resolve important open questions that have long remained out of reach, as we discuss in \cref{sec:open_dirs}.

\subsection{Why learning mechanics matters}

Building learning mechanics will not be easy.
It will require sustained effort, both intellectual and institutional.
It is therefore worth being clear about why such a project matters.
The reasons to seek a mechanics of learning fall into three broad categories: \textit{scientific}, \textit{practical}, and \textit{safety-related}.

The scientific reasons concern what such a theory could teach us about intelligence and the natural world.
The striking engineering success of large neural networks suggests that they exploit deep principles of learning and representation that we do not yet understand. 
This has historical precedent: technology has often preceded scientific theory, as was the case with steam engines' role in motivating thermodynamics, which went on to explain much more than engine efficiency.
A similar story played out in flight: the development of airplanes through trial and error and inspiration from the natural world helped motivate aerodynamic theory, which in turn enabled both better aircraft design and a deeper understanding of how birds themselves fly.
In our case, the principles that govern learning in artificial neural networks may also shed light on our own biological intelligence, with potentially important implications for neuroscience and cognitive science.

The practical reasons concern the design and development of real-world AI systems.
A mature theory of deep learning could guide model design, optimization, scaling, and deployment, replacing trial and error with more reliable principles.
For practitioners, concrete goals include better performance at fixed compute and data budgets, faster convergence, and less hyperparameter tuning.
Theory has already begun to support these goals through empirical scaling laws that guide compute allocation (\cref{sec:reason-measurable}), hyperparameter transfer from small to large models (\cref{sec:reason-hps}), and theoretically motivated optimizers and methods for data attribution (\cref{sec:reasons_for_skepticism}).
% Theory has already begun to play this role in a limited but growing number of cases, including empirical scaling laws (\cref{sec:reason-measurable}), mathematical prescriptions for hyperparameter scaling (\cref{sec:reason-hps}), and theoretically-motivated optimizers and methods for data attribution (\cref{sec:reasons_for_skepticism}).
% neural-tangent-kernel-based methods for data attribution \citep{park:2023-trak}, and theoretically motivated optimizers \citep{gupta2018shampoo,jordan2024muon}.
A deeper, more complete theory will give more such guidance and make it sharper and more predictive.

The safety reasons concern our ability to describe, characterize, and govern increasingly powerful AI systems. 
Some form of regulation will likely be necessary, but it is difficult to regulate a technology that we cannot clearly describe.
A theory that identifies the relevant variables, mechanisms, and organizing principles of large models could help provide the clarity needed for reliability, oversight, and control.
One avenue by which fundamental theory might aid in AI safety is by supporting mechanistic interpretability, a point to which we return in \cref{sec:learning_mech_and_mechinterp}.

\subsection{Plan for this paper}

This paper is structured as follows.
In \cref{sec:reasons_why}, we present our five lines of evidence that a scientific theory of deep learning is beginning to emerge.
We motivate each line of evidence with an intuitive explanation and highlight examples of research successes that illustrate the underlying principle.
In \cref{sec:learning_mech_and_mechinterp}, we discuss the relationship between learning mechanics and other perspectives on the science of deep learning, including a possible symbiotic relationship between learning mechanics and mechanistic interpretability.
In \cref{sec:reasons_for_skepticism}, we review and address common arguments that fundamental theory will not be possible.
In \cref{sec:open_dirs}, we give a portrait of ten important open directions in learning mechanics, from predicting scaling laws to eliminating hyperparameters, where we expect to see major progress in the coming years.
Finally, in \cref{sec:how_to_get_involved}, we offer some advice for young researchers looking to get involved in this scientific project and extend a hand with some introductory resources.

We write this paper for a broad audience.
We hope the \textit{veteran scientist of deep learning} will find something valuable in our synthesis of useful approaches and results, and feel galvanized by our depiction of an emerging science.
We hope to convince the \textit{deep learning practitioner} that theory is on a path to fulfilling its longstanding promise of practical utility and to encourage them to experiment with their systems with an eye for science.
We hope to convince the \textit{AI safety or mechanistic interpretability researcher} that white-box theory is difficult yet possible --- that a first-principles study of dynamics can help put solid foundations beneath their important work, and that our communities should work together (see \cref{sec:learning_mech_and_mechinterp} for our vision of symbiosis).
Lastly, we hope to make it easier for \textit{young students and newcomers to the field} to get involved.
This is an exciting and important area of work, and while it requires some mathematical maturity to get started, it is our belief that the barrier to entry could be much lower.
Various deep intuitions about this science have been percolating inside the theory community for a while, and this paper is an attempt to state them clearly.
We hope to make it easier for folks with the requisite background to quickly get up to speed and contribute.

\section{Evidence of an emerging mechanics of learning}
\label{sec:reasons_why}

A great cause for optimism that a mechanics of learning is possible is the fact that the essential ingredients of deep learning are both \textit{explicit} and \textit{measurable.}
A deep learning system is characterized by the following components:
\begin{equation*}
    \begin{aligned}
    \text{Architecture:} \quad & \text{a neural network } f(\vx;\vtheta) \text{ specified as a composition of simple linear and nonlinear transformations.} \\
    \text{Data:} \quad & \text{a dataset } \mathcal{D} = \{(\vx_i, \vy_i)\}_{i=1}^n \text{ consisting of samples from an unknown data-generating distribution }\\[-0.7ex]& (\vx,\vy) \sim \mathcal{P}_{\mathrm{data}}. \\
    \text{Task:} \quad & \text{an objective } \mathcal{L}(\vtheta) \text{ measuring the performance of the network } f(\vx;\vtheta) \text{ on the dataset } \mathcal{D}. \\
    \text{Learning rule:} \quad & \text{a gradient-based update equation, e.g. }
    \vtheta^{(t+1)} = \vtheta^{(t)} - \eta \nabla \mathcal{L}(\vtheta^{(t)})
    \text{, together with a parameter} \\[-0.7ex]
    & \text{initialization, e.g. } \vtheta^{(0)}_i \sim \mathcal{N}(0,\alpha^2_{\mathrm{init}}) \text{ and optimization hyperparameters, e.g. learning rate }\eta\text{.}
    \end{aligned}
\end{equation*}
Nothing about the learning process is hidden.
Unlike many complex systems where the equations governing dynamics must be inferred from observations, deep learning directly exposes its ``equations of motion.''
Moreover, these dynamics are extraordinarily measurable: every weight, activation, gradient, and loss value can be recorded, along with arbitrary statistics derived from them.
As a result, deep learning experiments are unusually easy to design, replicate, and interrogate, making it more straightforward to discover empirical regularities and rigorously test theoretical predictions.
Few fast-moving scientific domains offer comparable transparency in their governing equations or comparable freedom in what can be measured.

What, then, stands in the way of a scientific theory of deep learning? The central challenge is not \emph{opacity}, but \emph{complexity}.
While we have direct access to the architecture, data, task, and learning rule, the interaction of these components gives rise to learning dynamics that are nonlinear, coupled, and high-dimensional.
These dynamics depend in subtle ways on the choice of hyperparameters.
And even though we can inspect every training sample, data distributions are complex and have defied simple characterization.

Nevertheless, we argue that this complexity conceals underlying regularities, and that deep learning will indeed admit a scientific theory. 
In what follows, we present five broad observations that serve as evidence for an emerging mechanics of learning.
Each of these admits direct analogies to tools and ideas in other disciplines of mechanics.
These are summarized in \cref{tab:lm_physics_comparison}.

\newcommand{\celllist}[1]{%
  \parbox[t]{\linewidth}{\vspace{-2.3mm}#1}%
}

\begin{table}[H]
\centering
\renewcommand{\arraystretch}{1.25}
\setlength{\tabcolsep}{6pt}

\begin{tabular}{
    >{\centering\arraybackslash}p{0.07\textwidth}
    >{\RaggedRight\arraybackslash}p{0.21\textwidth}
    >{\RaggedRight\arraybackslash}p{0.29\textwidth}
    >{\RaggedRight\arraybackslash}p{0.31\textwidth}
}
\toprule
\textbf{Section} & \textbf{Approach} & \textbf{Examples in deep learning} & \textbf{Examples from physics%
% $^\dagger$
} \\
\midrule
\ref{sec:reason-dynamics}
&
solvable settings
&
\celllist{
deep linear networks,\\
kernel regression,\\
multi-index models
}
&
\celllist{
harmonic oscillator,\\
hydrogen atom,\\
Ising model
}
\\[3em]

\ref{sec:reason-limits}
&
simplifying limits
&
\celllist{
lazy vs.\ rich learning,\\
width, depth $\rightarrow \infty$,\\
small initialization
}
&
\celllist{
thermodynamic limit $(n, V \rightarrow \infty)$,\\
classical limit $(\hbar \rightarrow 0)$,\\
hydrodynamic limit $(\vk, \omega \rightarrow 0)$
}
\\[3em]

\ref{sec:reason-measurable}
&
simple empirical laws
&
\celllist{
neural scaling laws,\\
edge of stability,\\
neural feature ansatz
}
&
\celllist{
the laws of Kepler, Snell, Boyle, Hooke, Newton, Faraday, Ohm, Poiseuille, Planck, Hubble, etc.
}
\\[3em]

\ref{sec:reason-hps}
&
\celllist{
study of system\\
parameters
}
&
\celllist{
step size as sharpness\\
regularization,\\
$\mu$P and width-scaling
}
&
\celllist{
scaling analysis,\\
nondimensionalization,\\
chaotic vs. ordered regimes
}
\\[3em]

\ref{sec:reason-universal-behavior}
&
universal phenomena
&
\celllist{
common inductive biases and representations across models
}
&
\celllist{
critical phenomena,\\
renormalization group flow
% universal principles (stationary action\sout{Lagrangian minimization}, etc.)
% linear response phenomena
% ,\\ convergent evolution$^\dagger$
}
\\[2em]

\bottomrule
\end{tabular}

\caption{
\textbf{Useful tools and ideas in the emerging science of deep learning closely resemble important tools and ideas from physics, particularly classical mechanics, continuum mechanics, statistical mechanics, and quantum mechanics.}
Extrapolating, this suggests that there will be a \textit{mechanics of learning} which offers a unifying first-principles theory of the training process, hidden representations, final weights, and test-time performance of neural networks.
% $^\dagger$While convergent evolution comes from biology, not physics, we include it because the analogy is so strong.
}
\label{tab:lm_physics_comparison}
\end{table}

% Authors: Dan, Dhruva
\subsection{Analytically solvable settings exist}
\label{sec:reason-dynamics}

A reliable way to build scientific understanding in complex systems is to study pared-down yet representative settings in which quantitative calculations are possible.
For example, physics uses representative solvable settings like the harmonic oscillator and the hydrogen atom as sources of intuition for much broader classes of system.
Deep learning appears to be particularly amenable to this approach: scientists have identified a rich landscape of minimal models where the learning dynamics simplify and many quantities of interest become solvable.
These analytically tractable cornerstones are useful because they reveal phenomena and mechanisms to look for when we turn to realistic deep learning.%
\footnote{A complementary view is that any eventual complete theory of deep learning must encompass these simplified settings. Their solutions may provide conceptual scaffolding, serving as nucleation sites from which a more general theory crystallizes.}

One particularly fruitful simplification is \textit{linearization}.
Here we discuss two distinct instantiations of this idea: linearization in the data, where $f(\vx;\vtheta)$ becomes linear in $\vx$, and linearization in the parameters, where $f(\vx;\vtheta)$ becomes linear in $\vtheta$.
% 
% Here we discuss two distinct instantiations of this idea: linearization in the data and linearization in the parameters.

\begin{figure}
    \centering
    \begin{subfigure}{0.5\linewidth}
        \centering
        \includegraphics[width=\linewidth]{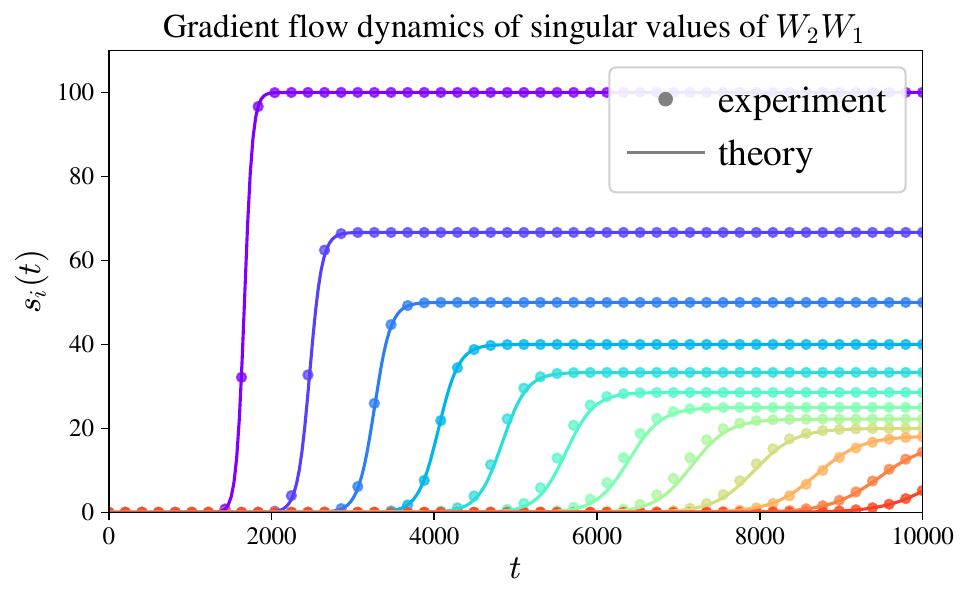}
        \caption{Linearization in the data}
        \label{fig:sub1}
    \end{subfigure}
    \hfill
    \begin{subfigure}{0.48\linewidth}
        \centering
        \includegraphics[width=\linewidth]{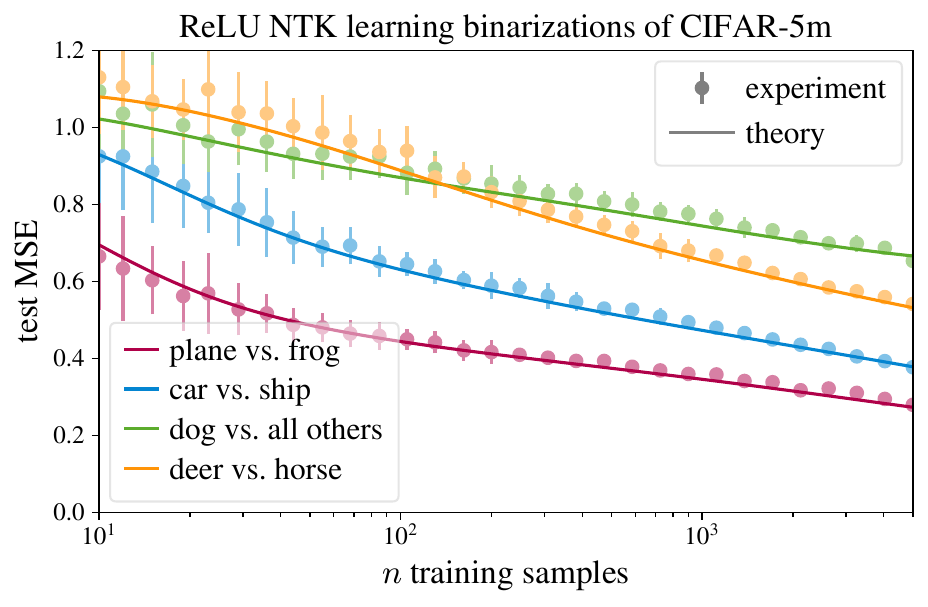}
        \caption{Linearization in the parameters}
        \label{fig:sub2}
    \end{subfigure}
    \caption{
    \textbf{Linearization yields exact solutions that match experiments.}
    (a) Canonical work by \citet{saxe2014exact} showed that, under a task-aligned initialization $\vtheta^{(0)}$ and whitened inputs $\vx \sim \mathcal{N}(0,\mathbf{I})$, the gradient flow learning dynamics of deep linear networks decouple into independent solvable Bernoulli ODEs.
    % \(e.g.\ $\dot{s}_i = s_i(s^\ast - s_i)$). 
    % 
    This leads to sequential learning of singular modes, with larger-singular-value modes emerging first. 
    Panel (a) reproduces Fig.~3 of \citet{saxe2014exact}.
    (b) Linearizing a nonlinear network by truncating nonlinear terms in its Taylor expansion around initialization reduces least-squares training to kernel ridge regression with the neural tangent kernel (NTK).
    This analysis connects the network's architecture to its inductive bias through the NTK eigenstructure, enabling accurate predictions for the test performance of these networks.
    Panel (b) is based on Fig.~2 of \citet{simon2023eigenlearning}.
    }
    \label{fig:linearization}
\end{figure}

\paragraph{Linearization in the data.}
A \emph{deep linear network} is obtained by removing all nonlinearities from a neural network's architecture, yielding a model that is linear in its inputs $\vx$ but remains highly \emph{nonlinear} in its parameters $\vtheta$:
\begin{equation}
    f(\vx;\vtheta) = \mathbf{W}_L \mathbf{W}_{L-1} \cdots \mathbf{W}_1 \vx,
    \qquad 
    \text{where }  \vtheta := \{\mathbf{W}_\ell\}_{\ell = 1}^L\text{, each }\mathbf{W}_\ell \text{ is a linear transformation, and } L \ge 2.
\end{equation}
Deep linear networks have a long history of study because, despite their simplicity, they retain many hallmark behaviors of deep learning \citep{nam2025position}.
These include saddle-point-dominated loss landscapes \citep{baldi1989neural}, dynamics with sharp phase transitions and separation of timescales \citep{gissin2019implicit,atanasov2021neural}, edge-of-stability oscillations with gradient descent \citep{even2023s}, and strong initialization-dependent inductive biases \citep{woodworth2020kernel, kunin2024get}.
Analysis of these networks is typically carried out with the \emph{gradient flow} learning rule --- the continuous-time limit of gradient descent --- under simplifying assumptions on the data distribution and with carefully chosen initializations \citep{fukumizu1998effect,saxe2014exact,tarmoun2021understanding,tian2021understanding,domine2025lazy}.
In these regimes, the learning dynamics can often be solved exactly or reduced to low-dimensional dynamical systems.

Across many such analyses, a consistent lesson emerges: learning exhibits a \emph{greedy} low-rank bias, acquiring some components of the task before others.
Canonical work by \citet{saxe2014exact} first showed how deep linear networks learn singular vectors of the input–output correlation sequentially during training, with learning prioritized toward modes associated with the largest singular values, as shown in \cref{fig:linearization}.
This bias has been hypothesized to benefit generalization by separating the signal from the noise \citep{lampinen2018analytic}, and closely mirrors behavior observed in nonlinear networks, where simpler functions are often learned before more complex ones \citep{kalimeris2019sgd,xu2019frequency,rahaman2019spectral,simon2023stepwise}.
Moreover, a range of factors --- including small initializations \citep{gidel2019implicit,li2021towards,jacot2021saddle,pesme2023saddle}, increased depth \citep{gunasekar2018implicit, arora2018optimization, arora2019implicit}, stronger mini-batch noise \citep{pesme2021implicit,chen2024stochastic}, and explicit $\ell_2$ regularization \citep{ziyin2022exact,wang2024implicit} --- have all been shown to further strengthen this greedy learning bias.

\paragraph{Linearization in the parameters.}

A \textit{linearized network} is obtained by truncating the nonlinear terms in a network's Taylor expansion around its initial parameters. This yields a model that is linear in its parameters $\vtheta$ but remains highly \emph{nonlinear} in the data $\vx$:
\begin{equation}
    f_\text{lin}(\vx;\vtheta) = f(\vx;\vtheta_0) + \nabla_{\vtheta} f(\vx;\vtheta_0)^\top (\vtheta-\vtheta_0),
    \qquad
    \text{where } \nabla_{\vtheta} f(\cdot;\vtheta_0)  \text{ is the gradient at initialization.}
\end{equation}
This is not some contrived construction: in fact, there are settings in which a model is well-approximated throughout training by its linearization, i.e., $\forall t,\  f(\vx;\vtheta_t)\approx f_\text{lin}(\vx;\vtheta_t)$.
% For example, a network in which only the final layer is trained exactly satisfies $f(\vx;\vtheta_t) = f_\text{lin}(\vx;\vtheta_t)$, where $\vtheta$ contains the last-layer weights \citep{lee2019wide}.
For example, any neural network architecture can be driven into the linearized regime by taking suitable limits \citep{jacot2018neural,lee2019wide,chizat2019lazy,liu2020linearity}, as discussed in \cref{sec:reason-limits}.
Additionally, recent evidence suggests that language model fine-tuning occurs in a near-linearized regime \citep{malladi2023kernel,ren2025learning}.

Since a linearized network is linear in its parameters, its learning dynamics are identical to those of linear regression, with one key difference:
while the dynamics of linear regression are driven by the Gram kernel, $K_\text{Gram}(\vx,\vx')=\vx^\top\vx'$, linearized networks are described by the \textit{neural tangent kernel} (NTK), $K_\text{NTK}(\vx,\vx')\coloneqq \nabla_{\vtheta} f(\vx;\vtheta_0)^\top \nabla_{\vtheta} f(\vx';\vtheta_0)$.
When the task is least squares regression and training uses small-step gradient descent, the dynamics are analytically tractable and the final predictor is given by kernel ridge regression with the NTK \citep{jacot2018neural}.

This setting yields insight into a variety of deep learning phenomena.
For example, since the details of the network architecture influence the mathematical structure of the NTK through the fixed feature map $\nabla_{\vtheta} f(\cdot;\vtheta_0)$, one learns how the linearized model's inductive bias follows from its architecture \citep{arora2019exact,geifman2020similarity}.
Furthermore, one may accurately predict the model's expected generalization error on arbitrary targets $f^\star$ by accounting for the structure of the input data \citep{jacot:2020-KARE,canatar2021spectral,loureiro:2021-learning-curves,hastie:2022-surprises,wei2022more,simon2023eigenlearning}, as shown in \cref{fig:linearization}.
Applying this framework to realistic data distributions uncovers the origin of the typical models' tendency to learn simple and generalizing functions \citep{basri2020frequency,karkada2025predicting}.
Linearized models also capture relevant phenomena such as double descent \citep{belkin2019reconciling, advani2020high} and scaling laws \citep{caponnetto2007optimal,pillaud2018statistical, cui2023error, atanasov2024scaling}.

However, despite these theoretical merits, linearized networks are unrealistic in a few critical ways.
Most notably, they do not capture the strong feature-learning capabilities that generic neural networks exhibit, often leading to overly pessimistic predictions for sample complexity \citep{ghorbani2020neural,vyas2022limitations}.
Moreover, by reducing training to a tractable linear problem, these models sidestep the intrinsically nonconvex optimization phenomena of deep learning.
To describe these and other aspects of deep learning, one must look beyond linearization.

\paragraph{Beyond linearization.}
An important frontier for theory lies in developing analytically tractable toy models that remain genuinely nonlinear in both the data and the parameters (see \cref{openquestion:models_of_nonlin_learning}).
In these settings, the influence of the data distribution becomes more complex, making it difficult to obtain a unified and general framework.
However, a growing body of work is progressing in this direction by isolating specific nonlinear mechanisms and making them solvable under assumptions on the data.

One line of work studies Gaussian inputs and structured targets (e.g., single- and multi-index models). Fully nonlinear neural networks provably outperform kernel methods using fewer samples because they exploit the structure in the target function to learn relevant features \citep{abbe2022merged, damian2022neural, bietti2022learning, ba2022high, dandi2023two}. Complementarily, methods from statistical physics enable computing exact asymptotics for Bayes-optimal inference and learning dynamics in these models \citep{barbier2019optimal, aubin2018committee, mignacco2020dynamical}. A related setting is two-layer neural networks with quadratic activation functions, where recent results have characterized the exact asymptotics, training dynamics, and scaling laws \citep{erba2025nuclear, arous2025learning, defilippis2025scaling, ren2025emergence}. Several other lines of research isolate distinct nonlinear phenomena: the convergence of homogeneous networks trained on logistic losses to max-margin solutions \citep{soudry2018implicit, lyu2020gradient}, the reduction of training dynamics to low-dimensional summary statistics in teacher-student models \citep{saad1995exact, tian2017analytical,goldt2019dynamics, ben2022high, veiga2022phase, zavatone2025summary}, memorization in associative memory models \citep{nichani2025understanding}, learned algorithmic structure in modular arithmetic tasks \citep{morwani2023feature, gromov2023grokking, tian2024composing, kunin2025alternating}, nonlinear solvable models of attention \citep{tian2023scan, tian2024joma, zhang2025training, boncoraglio2025singlehead}, feature learning in nonlinear contrastive models \citep{tian2022coordinate,tian2023nonlinearity},
% hierarchically structured targets \citep{cagnetta2024deep} (\dhruva{is the hierarchy model an example of a solvable model? afaict the dynamics are not solved in this paper.}),
and improved scaling laws from nonlinear feature learning \citep{bordelon2025feature}.

Taken together, these approaches illustrate both the promise and the limitations of current nonlinear toy models: each captures a slice of fully nonlinear learning dynamics, yet no unified framework has emerged.
We view this space as an open and rapidly evolving area, and return to these challenges in our discussion of open problems in \cref{sec:open_dirs}.

\subsection{Insightful limits reveal fundamental behavior}
\label{sec:reason-limits}

Modern deep learning systems are enormous: they regularly involve hundreds of interacting architectural components comprised of hundreds of billions of parameters and trained on trillions of tokens.
With so many interacting degrees of freedom, constructing detailed microscopic theories that track individual parameters in practical systems seems all but hopeless.

Fortunately, complex systems often simplify when approximated as effectively \emph{infinite} in size, revealing simple mathematical structure that remains informative even for the original finite system.
This strategy is well established in statistical and chemical physics: for example, the ideal gas law, $PV = nRT$, is derived in a limit of infinite number of particles (often termed the \textit{thermodynamic} limit) yet accurately describes real parcels of gas of finite volume.
Limits are a central mathematical tool for managing the complexity of deep learning, and their recurring success in doing so provides strong evidence for an emerging theory.

Here we discuss the \textit{limit of infinite width} in detail.
We conclude by mentioning other limits and offering some unifying ideas.

\paragraph{The infinite width limit and the lazy/rich dichotomy.}
The dynamics of a deep neural network often simplify when one takes the number of neurons in each hidden layer to infinity.
Such a limit generally leads to so-called \textit{mean-field} behavior in which we only need to describe the evolution of the neuron population as a whole (as e.g. a probability distribution) and we can ignore what each individual neuron is doing.
However, achieving this limit requires shrinking the initialization scale as width increases to prevent activations in deeper layers from diverging.
The key subtlety in taking the infinite width limit is that the \textit{rate} at which we suppress these initial weights strongly influences the resulting training dynamics, leading to one of two qualitatively distinct limiting behaviors.

\textit{The lazy, kernel, or linearized regime.}
The first forays into the land of infinite width studied only a network's statistics at initialization, not its training dynamics \citep{neal:1996,poole:2016}.
These works found that, in order for the inputs to hidden neurons to neither vanish nor explode as width increases, the parameter size at initialization has to decay as $\text{[width]}^{-1/2}$.
This is not a surprise: it is just the well-known LeCun initialization rule \citep{lecun:1998}, which can be easily derived from the central limit theorem.
Later works that tried naively training the parameters of these infinite-width networks found the surprising fact that the network's \textit{weights and hidden representations} change only negligibly, yet these small changes accumulate to produce substantial changes in the output function.
As a result, the training dynamics are \textit{linear in the parameters} in the sense discussed in \cref{sec:reason-dynamics}, and the evolution of the target function may be expressed entirely in terms of the NTK \citep{jacot2018neural,lee2019wide}.
While a network in this limit is wonderfully analytically tractable, the fact that its hidden representations evolve only negligibly means that it fails to exhibit \textit{feature learning}.
While the definition of feature learning is much debated (see \cref{openquestion:what_are_features}), all agree that at minimum it requires the network's hidden activations on a given data sample to change from their values at initialization, which does not happen in this limit.
This suggests the NTK infinite-width limit is not the right one to study.
Networks in this linearized regime were later termed ``lazy'' by \citet{chizat2019lazy}.

\begin{figure}
    \centering
    \includegraphics[width=15cm]{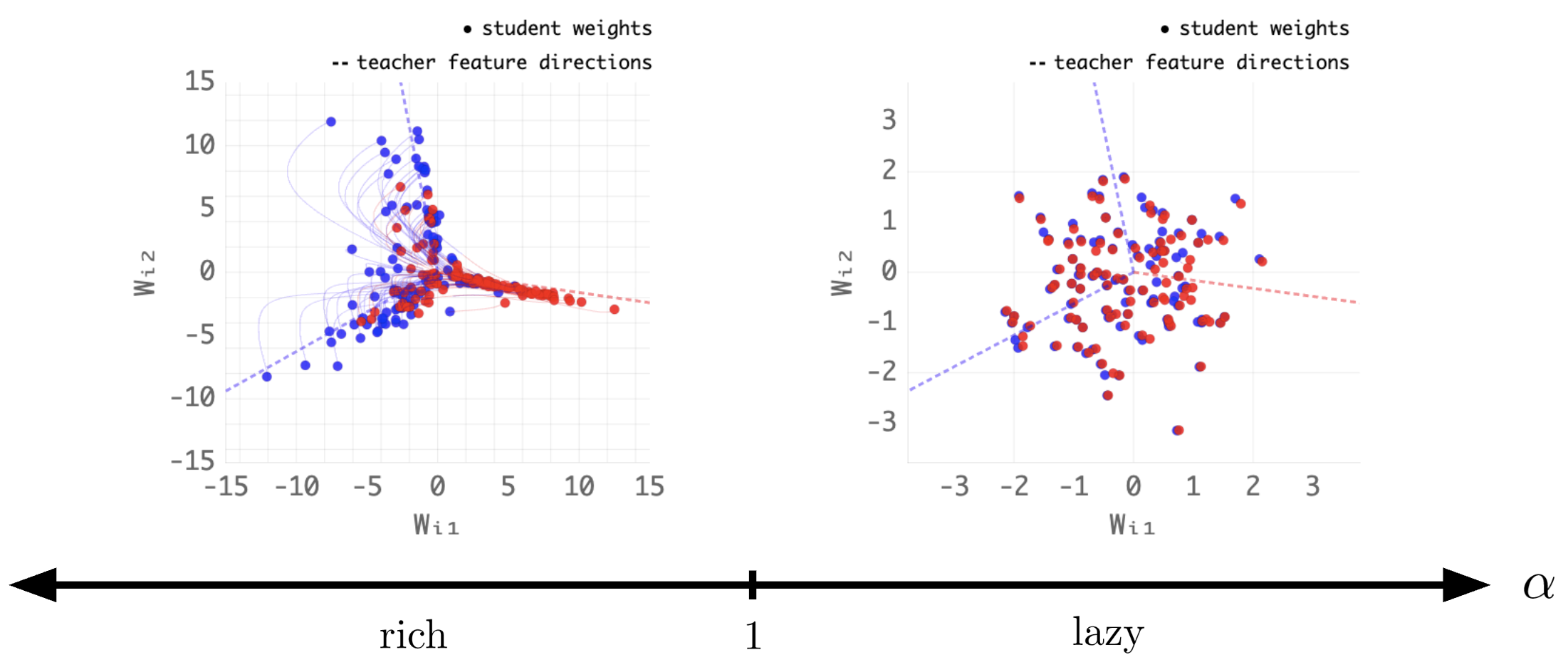}
    \caption{
    \textbf{Large and small network output multipliers are sufficient to induce lazy and rich training dynamics.}
    We train a shallow student network $\hat{f}(\vx) = \frac{\alpha}{n} \sum_{i=1}^n a_i \text{ReLU}(\vw_i^\top \vx)$ with width $n = 200$ to match a teacher network $f^*(\vx) = \sum_{i=1}^3 a^*_i \text{ReLU}((\vw^*_i)^\top \vx)$ on two-dimensional input data.
    We plot the training trajectories of the student weights $w_i$ (color denotes $\mathrm{sgn}(a_i)$) against the teacher feature directions.
    \textbf{Left:} with $\alpha = 0.1$, the dynamics are \textit{rich:} the student weights grow significantly and cluster in angle around the teacher feature directions.
    \textbf{Right: } with $\alpha = 30$, the dynamics are \textit{lazy:} the student weights move negligibly during training, even though the loss drops.
    Experiment reproduced from \citet{chizat2019lazy}.
    }
    \label{fig:lazy_vs_rich}
\end{figure}

\textit{The rich, active, or feature-learning regime.}
In answer to this, several authors identified an alternative infinite width limit in which training \textit{does} induce feature learning.
The key insight was essentially to downscale the final-layer weights by a factor of $\text{[width]}^{-1}$, rather than the earlier $\text{[width]}^{-1/2}$, thereby forcing the network weights to change more to compensate.%
\footnote{The lazy vs. rich dichotomy is conceptually similar to elastic vs. plastic deformation in materials. A material will deform linearly in response to a small force, and its internal atomic structure will not change. In response to a larger force, it will deform nonlinearly, and its internal structure changes.}
While this makes the function trivial at initialization (at infinite width it is uniformly zero), it can still grow nontrivially during training, changing by an order-one amount upon each gradient step.

This downscale-the-network-output idea first appeared in the shallow ``mean-field networks'' of \citet{mei2019mean}, \citet{rotskoff2018parameters}, and \citet{chizat2018global}.
\citet{geiger2020disentangling} and \citet{yang2021tensor} found that this idea also works for networks of arbitrary depth, bundling the resulting hyperparameter scaling factors together into the celebrated ``Maximal Update Parameterization'' discussed in \cref{sec:reason-hps}.
It is now widely accepted that infinite-width neural networks can learn features.

Wide networks in this ``rich'' regime display a huge range of interesting behaviors that their lazy counterparts do not.
The most significant is certainly that the hidden features of these networks change over time, adapting to the structure in the input data, altering the internal geometry of hidden representations over the course of training \citep{bordelon2022self}.
Subpopulations of neurons specialize, learning to attend to different features latent in the data \citep{aubin2018committee, goldt2019dynamics, ren2025emergence}. 
For instance, in tasks where the optimal predictions involve low-dimensional subspaces of high dimensional data, the distribution over first layer weights evolves to amplify weights in the subspace of interest \citep{mei2018mean, abbe2022merged, moniri2023theory, cui2024asymptotics, defilippis2025scaling, erba2025nuclear, montanari2026phase}.
When the scale of the initialization is made even smaller, they often show the greedy low-rank bias discussed in \cref{sec:reason-dynamics}, acquiring some components of the task before others \citep{saxe2015deep,atanasov2021neural,atanasovoptimization}.%
\footnote{
There is also a well-developed line of work studying the signatures of feature learning in large-width networks from a \textit{Bayesian} perspective.
Naively, infinite-width networks have simple Bayesian statistics given by Gaussian processes \citep{lee2017deep}, which is analogous to the ``lazy'' limit of conventionally-trained networks.
This view treats this Gaussian process limit as a solvable reference point \citep{cohen2021learning,lavie2024towards} and then reintroduces finite width, using mean-field and variational techniques to characterize aspects of feature adaptation to data \citep{cohen2021learning,seroussi2023separation,rubin2023grokking,pacelli2023statistical,rubin2025kernels,rubin2025mitigating}.
One may also induce feature learning by rescaling the total likelihood (see e.g. \citet{yang2023theory}), which is analogous to the final-layer downscaling which gives the rich limit in conventional training.
}

The lazy–rich dichotomy, and its dependence on initialization scale, emerged as a central finding of infinite-width analyses. 
Subsequent work has shown that analogous behavior appears even at finite width: scaling down the network output promotes feature learning, pushing models toward the rich regime, whereas increasing the output scale tends to linearize training dynamics and induce lazy behavior \citep{chizat2019lazy}.
This sensitivity to initialization scale connects to a broader literature on inductive bias, where seemingly small changes to the learning setup can steer training toward fundamentally different solution classes \citep{maennel2018gradient,woodworth2020kernel,luo2021phase}. 
\cref{fig:lazy_vs_rich} illustrates how the same finite network, trained with different output scalings, can exhibit either lazy or rich learning dynamics.

\paragraph{The infinite depth limit and other hyperparameter limits.}

As with infinite width, one can arrive at a stable infinite depth limit of a deep residual network by downscaling the contribution of each layer so the residual stream does not blow up.
Here, too, there are different limiting behaviors depending on the size of this downscaling factor: suppressing each layer by a factor of $\text{[depth]}^{-1}$ results in limiting dynamics in which the residual stream changes smoothly over depth \citep{bordelon2024infinite, chizat2025hidden, chaintron2026resnets} (reminiscent of Neural ODEs \citep{chen2018neural}) while suppressing each layer by a factor of $\text{[depth]}^{-1/2}$ results in limiting dynamics in which the residual stream diffuses as if driven by a stochastic differential equation \citep{bordelon2023depthwise, yang2023depth}.
Networks in these two limits converge to qualitatively different solutions in realistic architectures such as transformers \citep{dey2025don}.
It is not yet clear which is the more important limit to study.

Some deep learning architectures admit size limits other than those of large width or large depth. Instead of increasing size or total number of distinct feedforward layers, one can also analyze the infinite limits of recurrent architectures using similar mean-field ideas \citep{clark2026structure, bauer2026unified}. State-of-the-art transformer models include more expressive constituent blocks such as multi-head self-attention layers and mixture-of-expert multi-layer perceptrons. These layers have multiple scaling directions including head count, head size, and context length for attention \citep{hron2020infinite,bordelon2024infinite} and expert count, expert size, and sparsity for mixture-of-expert models \citep{pioromu2025moes,jiang2026hyperparameter}.
Clarifying the interplay of different infinite limits in these models is important to making contact with modern practice and to disentangle various hyperparameters related to initialization and optimization (see Section \ref{sec:reason-hps}). 

Lastly, most optimization hyperparameters have an associated limit.
As the batch size approaches infinity, we reach population gradient descent.
% (and start to see diminishing returns from increasing batch size further \citep{mccandlish2018empirical}).
% but we also encounter the critical batch size \citep{mccandlish2018empirical}.
As we take learning rate to zero, we recover gradient flow.
If we add an infinitesimal weight decay and take training time to infinity, we first optimize the loss to convergence, then perform parameter norm minimization conditioned on the final value of the loss.
% \dan{@Jamie -- do we want to cite here equilibrium/bayesian perspective used by Zohar and Haim -- although its not vanishing weight decay.}
We discuss how to understand the corrections induced by having finite values for some of these hyperparameters in \cref{sec:reason-hps}.

\paragraph{Joint scaling limits.}

Sometimes scaling limits in multiple variables ($\nu_1,\nu_2$) play nicely, in the sense that $\displaystyle \lim_{\nu_1 \to \infty}\,\lim_{\nu_2 \to \infty}$ gives the same result as
$\displaystyle \lim_{\nu_2 \to \infty}\,\lim_{\nu_1 \to \infty}$.
For example, the infinite width and depth limits in residual networks usually commute in this way, so long as one takes a sensible parameterization \citep{hayou2023width}.
However, in many theoretical machine learning settings, different scaling dimensions do not commute, and the limiting behavior could depend on a limiting ratio $\nu_2 / \nu_1$.
Such joint/proportional scaling limits are common in random matrix theory: for example, consider the SVD of a random matrix with $P$ rows and $N$ columns with $N,P \to \infty$ with $P/N$ held constant.
In machine learning theory, neural networks trained with random data can often be described by a joint scaling limit where both the dataset size and parameter count are taken to infinity, but one or more of the \textit{ratios} 
% $\text{[data]} / \text{[parameters]}$
$\frac{\text{[data]}}{\text{[input dim]}}$, $\frac{\text{[data]}}{\text{[width]}}$, or $\frac{\text{[data]}}{\text{[parameters]}}$
is a finite value \citep{seung1992statistical,saad1995exact,zdeborova2016statistical,li2021statistical,maillard2024bayes,martin2024impact,barbier2025statistical}.
This joint scaling is likely necessary in the study of \textit{compute-optimal neural scaling laws} where the training horizon (i.e. dataset size) is scaled linearly with the total parameters \citep{hoffmann:2022-chinchilla} and to theoretically characterize hyperparameter transfer phenomena \citep{bordelon2025deep}. These joint (data \& model size) limits are potentially important as infinite parameter limits at fixed dataset size are capable of perfect interpolation and do not capture scaling law behaviors across model sizes (see Section \ref{sec:reason-measurable}). 
Other well-studied joint scaling quantities include the ratio
% $\text{[width]} / \text{[depth]}$
$\frac{\text{[width]}}{\text{[depth]}}$
in non-residual networks \citep{hanin2019finite, li2022neural, noci2023shaped, hanin2025global}, the ratio
% $\text{[learning rate]} / \text{[output multiplier]}$
$\frac{\text{[learning rate]}}{\text{[output multiplier]}}$
in the rich regime \citep{atanasovoptimization}, and the ``SGD noise temperature''
% $\text{[learning rate]} / \text{[batch size]}$ 
$\frac{\text{[learning rate]}}{\text{[batch size]}}$ \citep{mandt2017stochastic,jastrzkebski2017three}.
% \florentin{It could be worth mentioning here that in RMT, whether N and P limits commute depend on the assumptions on the limit of the covariance: if it is trace-class (described by a decreasing sequence of eigenvalues rather than a measure) then they do commute!}

\paragraph{The Discretization Hypothesis.}
Overall, the widespread use of limits to manage the complexity of deep learning reflects a recurring theme across scientific disciplines: appropriate asymptotic perspectives often render otherwise intractable systems analytically tractable.
Many theorists share a heuristic belief that most practical neural networks can be understood as noisy, finite approximations to models of infinite size.%
\footnote{Works studying finite-size corrections to infinite limits include \citep{hanin2019finite,roberts2022principles, zavatone2021asymptotics, segadlo2022unified,bordelon2023dynamics, glasgow2025propagation}.}
By analogy, one numerically solves a partial differential equation by discretizing over space and time, and the finer the discretization, the smaller the numerical error from the desired continuum process.
This is very possibly also true of deep neural networks, with width and depth taking the place of space and time.
Other finite hyperparameters, such as the learning rate, batch size, and dataset size, might also be understood in this way.

We might call this belief the \textit{Discretization Hypothesis.}
While it has yet to be made precise or proven (see \cref{openquestion:discretization}), this hypothesis has implicitly underpinned much important work, and little in the analytical study of large models makes sense without it. 
% \jeremy{i think this wording might go too far} \florentin{What would falsify this hypothesis exactly? The absence of a limit, or the fact that the convergence to it is impractically slow? Or is there more to it than this?}

The Discretization Hypothesis amounts to the statement that finite-size corrections from limits typically worsen performance while saving costs in data, time, memory, and compute.
Showing that these finite-size effects deliver a general benefit that cannot be achieved any other way would falsify this hypothesis.

\subsection{Simple empirical laws capture meaningful macroscopic statistics}
\label{sec:reason-measurable} 

% Even when first-principles analyses via solvable models (\cref{sec:reason-dynamics}) or asymptotic limits (\cref{sec:reason-limits}) are unavailable, deep learning’s incredible measurability makes it possible to discover simple, predictive relations at the level of macroscopic statistics.

Deep learning is highly measurable: it is easy to track a vast array of quantities before, during, and after training.
While any quantity \textit{can} be measured, the most lawful are typically \textit{aggregate, macroscopic statistics} over many weights and samples.
For instance, the train and test losses are aggregates over many samples.
These quantities are occasionally described by simple empirical laws relating one to another.
Such laws have already played an important role in shaping both our understanding and practice of deep learning.

This pattern has ample precedent in the quantitative sciences. 
Many important physical and chemical laws were first discovered as empirical regularities and only later understood in terms of deeper principles, including laws due to Kepler, Snell, Boyle, Hooke, Faraday, Ohm, Poiseuille, and Planck.
Given how often scientific fields have developed in this way, it seems likely that deep learning will continue to yield empirical laws as its science matures.
% 
% \joey{We should, however, be cautious to prescribe laws to useful macroscopic statistics.
% %
% For example, the NTK alone completely describes linearized networks \cref{sec:reason-dynamics}, but for most feature-learning networks, there are no simple equations describing the NTK.
% }
% %
Here, we highlight a handful of examples %\joey{of lawful macroscopic statistics}
and conclude with takeaways for theorists.

\begin{figure}[ht]
  \centering
  \includegraphics[width=\linewidth]{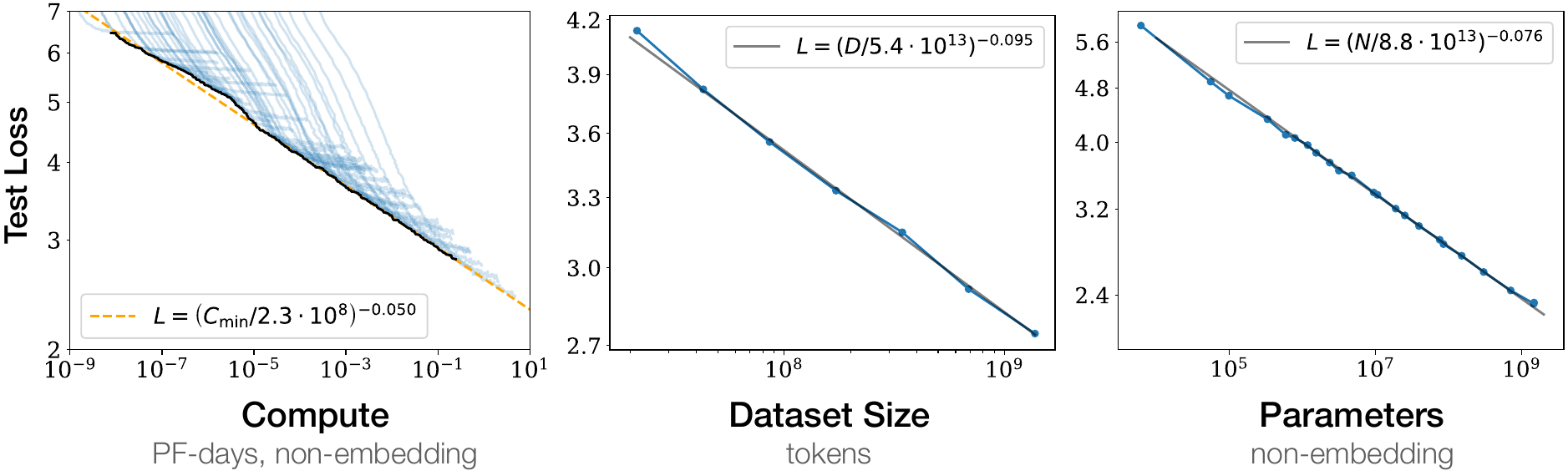}

\caption{\textbf{The loss of large neural networks decays according to predictable \textit{neural scaling laws}.}
These neural scaling laws take the form of power laws (linear on log-log plots) in compute, dataset size, and parameter count.
Adapted from \citep{neuralscalinglaws}.
}
\label{fig:scaling_laws}
\end{figure}

\paragraph{Neural scaling laws.} 

The single most important measurement of any machine learning system is its test loss.
Given the complexity of large deep learning systems, one might expect the test loss to be a complex, unknowable function of the system's hyperparameters.
This is not so: studies of \textit{neural scaling laws} \citep{neuralscalinglaws, neuralscalinglawsearly} demonstrate that, within an architectural family, the final loss follows a predictable power law function governed by only three scalar variables: compute, the amount of data, and the network's size.
% Neural scaling laws are not prescriptive in the same way Newton's laws of motion define how objects must move, but are an average-case, descriptive law  like those found in thermodynamics.
These power laws are shown in \cref{fig:scaling_laws}.

Why does test loss decay as a power law in these variables, and what determines the scaling law exponent?
We still do not know.
While scaling laws are often attributed to structure in the data, with candidate explanations in terms of the dimensionality of the data manifold \citep{sharma2022scaling,neuralscalinglawsexplain}, feature superposition \citep{liu2025superpositionyieldsrobustneural}, and power laws latent in task structure \citep{cui:2021-krr-decay-rates,bordelon:2024-model-of-scaling-laws,michaud2023quantization, ren2025emergence, defilippis2025scaling}, they may also depend on details of the architecture and optimizer \citep{barkeshli2026origin,liu2026universal}.
At present, no framework can robustly predict the observed exponents \textit{a priori} from dataset and architectural properties across realistic settings (see \cref{openquestion:predict_powerlaw_exponents}), though recent progress has begun to move in this direction \citep{cagnetta2026deriving}.
The fact that test loss is so predictable strongly suggests that a simple underlying explanation remains to be found.

\paragraph{Weight dynamics at the edge of stability.}
Because every model is the result of a training process, we would like to understand the dynamics and trajectory of a model's weights during training.
While there are simple cases where these dynamics are exactly solvable (see \cref{sec:reason-dynamics}), this is usually well out of reach.
The loss landscape dictates the network's dynamics, but a direct visualization of the loss, as is done in \citet{losslandscapevis}, suggests an immensely complicated landscape that is unlikely to have lawful regularities.

\begin{figure}[ht]
  \centering
  \includegraphics[width=\linewidth]{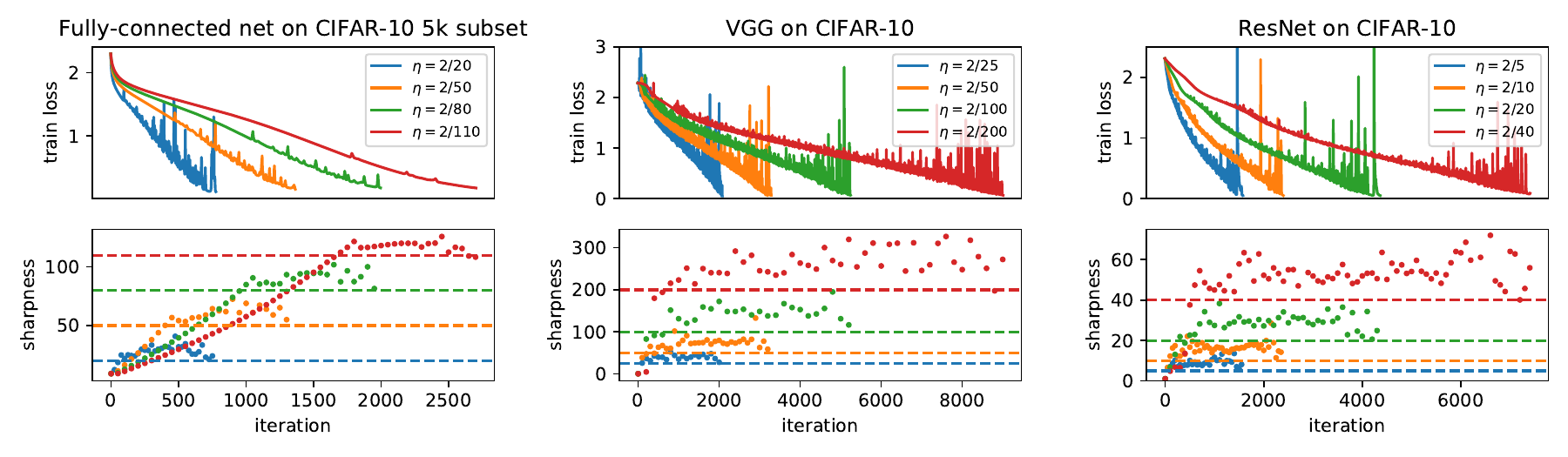}

  \caption{
  \textbf{Gradient descent occurs near the edge of stability.}
    Three architectures are trained with full-batch gradient descent on CIFAR-10 with varying learning rate $\eta$.
    Plots show the train loss (top row) and Hessian sharpness (bottom row).
    For each step size $\eta$, observe that the sharpness rises to $2/\eta$ (dashed horizontal lines) and hovers at or just above this value.
    Adapted from \citet{cohen2021gradient}.
  }
  \label{fig:measurable_eos}
\end{figure}

Nonetheless, some robust patterns in the coarse, aggregate properties of weight trajectories have been found.
One of these is the \textit{sharpness} of the network loss surface, defined as the largest eigenvalue of the Hessian with respect to the parameters.\footnote{Studies of the broader Hessian spectrum have linked spectral outliers to class structure \citep{papyan2019measurements,papyan2020traces}. Related analyses of trained weight matrices reveal heavy-tailed spectra and statistics that correlate with model performance \citep{martin2021implicit,martin2021predicting}.}
When a typical network is trained using full-batch gradient descent with learning rate $\eta$, the sharpness undergoes two distinct phases: a gradual increase (termed \textit{progressive sharpening}) followed by a plateau near $2 / \eta$, a regime termed the \textit{edge of stability} \citep{cohen2021gradient} (see \cref{fig:measurable_eos} and \citet{wu2018sgd} for an earlier sign of this behavior).

Having identified these regularities, we can begin to understand them.
Progressive sharpening provably occurs in deep linear networks \citep{even2023s,yoo2025understanding}, yet a quantitative explanation suitable to realistic nonlinear networks remains to be found (see \cref{openquestion:loss-curvature}).
More is understood about why the sharpness stabilizes at $2 / \eta$.
This value is the stability boundary for gradient descent on a quadratic loss: along a Hessian eigendirection with curvature greater than $2/\eta$, the dynamics exhibit oscillations of increasing magnitude.
Nonquadratic loss geometry can prevent this divergence, as earlier studies demonstrated through the catapult mechanism \citep{lewkowycz2020large}, subquadratic loss growth \citep{ma2022beyond}, and oscillation-induced motion toward flatter regions \citep{arora2022understanding,lyu2022understanding}.
\citet{damian2022self} developed a more general quantitative account of this \textit{self-stabilization}: third-order loss derivatives couple unstable oscillations to motion that reduces sharpness, counteracting progressive sharpening and keeping sharpness near $2/\eta$.
% Particularly, $2 / \eta$ is the maximum stable sharpness achieved in convex optimization--any sharpness larger than $2/\eta$ would cause parameter oscillations of increasing magnitude.
% In more general cases, \citet{damian2022self} showed how coarse properties of the \textit{third-order} loss curvature can cause the (second-order) sharpness to stabilize at $2 / \eta$.
Follow-up work reveals that loss dynamics at the edge of stability can be decomposed as \textit{smooth, time-averaged, gradient flow dynamics} plus \textit{oscillations in unstable directions} \citep{centralflows}.
These works make quantitative predictions about the parameter trajectory which closely match experiment.

\paragraph{Coarse properties of hidden representations and weights.}

There are a handful of other cases in which coarse properties of neural networks' hidden representations and weights are known to obey simple equations.
We will briefly mention three of these.

\textit{Neural collapse.}
Consider a neural network classifier trained to choose among $C$ classes.
\citet{neuralCollapse} found that, at the end of training, the final-hidden-layer representations of samples from each class tend to cluster tightly around their class mean.
Furthermore, the $C$ class mean vectors form a regular simplex.
Later theoretical work has explained this geometric arrangement as the natural energy-minimizing configuration when (a) the loss used is cross-entropy and (b) a small amount of weight decay is applied \citep{zhu2021geometric}.\footnote{This parallels how gradient descent on separable logistic regression converges in direction to the max-margin separator \citep{soudry:2018-implicit-bias-of-gd}. This connection between max-margin bias and neural collapse is made explicit in \citet{kunin2023asymmetric}.}
Neural collapse also has a layerwise counterpart: class separation improves progressively with depth \citep{zarka2020separation,rangamani2023feature} and often in a regular fashion \citep{he2023law}.\footnote{A related empirical law has also been observed in large language models: each layer reduces a measure of next-token prediction error by approximately the same factor \citep{he2025law}.}\citet{shi2025spring} propose a phenomenological model which reproduces qualitative features of this progressive separation.

\textit{The neural feature ansatz.}
At the other end of the network, there are some robust regularities known about the \textit{first}-layer weights.
\citet{neuralFeatureAnsatz} show that, after training, the Gram matrix of the first-layer weights $\mW_1^\top \mW_1$ aligns with the \textit{average gradient outer product}:
\begin{equation}
    \mW_1^\top \mW_1
    \propto
    \E{\vx \sim \mathcal{P}_\text{data}}{
    \nabla_{\vx} f(\vx; \vtheta)
    \nabla_{\vx} f(\vx; \vtheta)^\top
    },
\end{equation}
% where $f(\vx)$ is the network's feedforward function.
where $\nabla_{\vx} f(\vx; \vtheta)$ denotes the Jacobian of the network with respect to $\vx$.
While this rule is heuristic and inexact, it often makes strikingly accurate predictions for quantities like the top eigenvectors of $\mW_1^\top \mW_1$.
Similar heuristics hold at deeper layers.
At time of writing, there are only partial theoretical explanations for this phenomenon; see \citet{ziyin2024formation,featuresAtConvergenceTheorem}.

\textit{Gradient flow conservation laws.} 
A striking regularity identified in linear networks is that the difference between the covariance and Gram matrices of consecutive layers $\mW_\ell\mW_\ell^\top - \mW_{\ell + 1}^\top\mW_{\ell + 1}$ is conserved under gradient flow \citep{saxe2014exact,du2018algorithmic, arora2019convergenceanalysisgradientdescent}.
What initially appeared to be a curiosity of linear networks was later shown to follow from continuous symmetries of the parameterization --- an instance of the Noether principle --- and thus could be used to identify similar conserved quantities in nonlinear networks \citep{kunin2021neural, tanaka2021noether, marcotte2024abidelawfollowflow, marcotte2024momentumconservationlawseuclidean}.
For instance, the rescaling symmetries in networks with homogeneous nonlinearities (e.g., ReLU), the scale symmetries preceding normalization layers (e.g., batch normalization), the translation symmetries in the logits preceding a softmax, and the rotation symmetries between key and query matrices in attention all lead to symmetry-specific statistics of the parameters that are conserved under gradient flow and weakly broken by SGD in predictable ways.

\paragraph{Takeaways for theorists.} Theory can be built ``bottom-up,'' starting from first-principles math as in \cref{sec:reason-dynamics,sec:reason-limits}, or ``top-down,'' starting from empirical observations and attempting to explain them.
In this section we have highlighted a few notable examples of top-down theories.
We expect more to come.
The measurability of deep learning makes observation and empiricism a particularly fruitful approach, since experimentation can be iterated on quickly, while revealing mathematically simple relations and structure in trained models.
Of course, some caution is necessary: most macroscopic statistics \textit{don't} obey a simple and general mathematical law --- or at least don't seem to until plotted against the right quantity --- and so the challenge is to find those that do.
We encourage theorists of deep learning to proactively use experiments to look for lawful regularities in neural networks.

\subsection{Hyperparameters can be disentangled and understood}
\label{sec:reason-hps}

Training a deep learning system involves many numerical knobs, termed ``hyperparameters.'' These include optimization hyperparameters such as the learning rate, batch size, momentum, and initialization variance, as well as architecture hyperparameters such as width and depth. 
The large number of hyperparameters in deep learning presents a challenge not only for practitioners, who must tune them carefully in order to achieve optimal performance, but also for researchers, who must grapple with many confounding factors when trying to interpret the outcome of scientific experiments.
It is only in the last few years that the theory community has come to realize that hyperparameters can be disentangled and understood, and that the resulting mathematics is often both useful for practitioners and clarifying for theorists.

% [added this bc we want to discuss the stuff in \cref{tab:lm_physics_comparison}]
%
This study of hyperparameters bears similarities to the study of the constant parameters governing the behavior of a physical dynamical system.
For example, in a fluid flowing through a pipe, a dimensionless number called the \textit{Reynolds number} computed from the pipe diameter and the fluid's speed, density, and viscosity determines whether flow is laminar or turbulent.
While \textit{solving} for the trajectory of the turbulent fluid is extremely difficult, it is nonetheless very helpful to be able to quickly predict whether flow will be turbulent at all --- and how things change if you scale up the pipe diameter or increase the fluid flow.
Similarly, while solving the optimization dynamics of a neural network is very difficult, it is often very helpful to quickly obtain a coarse picture of how things change if you change one or more hyperparameters.
In this section we highlight two lines of work in which hyperparameters have been found to admit explanatory theory.

\paragraph{Understanding optimization hyperparameters.}

Stochastic gradient descent has two hyperparameters: learning rate and batch size.  The algorithm's dynamics are often invariant under a simultaneous rescaling of both.  That is, if one doubles both the learning rate and batch size, and halves the number of optimizer steps (or equivalently, keeps fixed the number of training examples processed), then the trajectory stays nearly the same.  This so-called \emph{linear scaling rule} \citep{goyal2017accurate} is useful for transferring a learning rate that was tuned for one batch size to a different one.  
% A line of theoretical work has shed light on this rule of thumb by interpreting SGD as a discretization of an underlying stochastic differential equation (SDE) \citep{jastrzkebski2017three,li2021validitymodelingsgdstochastic}, a model which implies the linear scaling rule.  
A line of theoretical work has clarified this rule of thumb by interpreting SGD as a discretization of an underlying stochastic differential equation (SDE), a perspective that predicts the linear scaling rule \citep{mandt2017stochastic, jastrzkebski2017three, chaudhari2018stochastic, li2019stochastic, li2021validitymodelingsgdstochastic}.
\citet{malladi2022sde} extended this line of work from SGD to adaptive optimizers, for which they argued that the learning rate should scale with the square root of the batch size.

% When choosing the batch size to train at, there is an inherent tradeoff between the two resources of serial time and overall compute (often equivalent to cost) \citep{ma2018power,jain2018parallelizing,mccandlish2018empirical, shallue2019measuring}.  
This invariance perspective explains how to adjust hyperparameters across batch sizes, but not how to choose the batch size itself. That choice involves an inherent tradeoff between two resources: serial time (the number of sequential training steps) and overall compute (the total amount of computation, often closely tied to cost) \citep{ma2018power,jain2018parallelizing,mccandlish2018empirical, shallue2019measuring}.
For a practitioner who cares only about serial time and not at all about cost, the optimal batch size is the full dataset.  Conversely, for a practitioner who cares only about cost and not at all about serial time, the optimal batch size is 1.  In reality, no practitioner falls exactly in either bucket; a practitioner might care more about one resource than the other, but is generally willing to accept some slack in return for a better deal on the second resource. A frequently discussed concept is that of the \emph{critical batch size}, a batch size which trades off between these two concerns.
% \cite{mccandlish2018empirical} presented a model for reasoning about this tradeoff between time and cost, under which the Pareto frontier of serial time vs. cost has the form of a hyperbola.
\citet{mccandlish2018empirical} proposed a simple model of this tradeoff under which the Pareto frontier between serial time and compute takes the form of a hyperbola.

Optimization hyperparameters in deep learning affect not just the speed and cost of training but also the \textit{trajectory} that training follows \citep{wen2024understanding}.
This in turn affects various properties of the learned network, including generalization performance \citep{keskar2016large,schulman2025lora} and compressibility \citep{catalan2025training, barsbey2025large}.
A fruitful line of work has sought to explain these effects through the hypothesis that many implicit effects of optimizer hyperparameters can be understood as \emph{implicit regularization of loss function curvature}.\footnote{An additional, but apparently weaker, effect is captured by an implicit regularization of the gradient norm \citep{barrett2020implicit, smith2021origin}.}
Empirical studies initially observed that first-order optimizers regularize the curvature (i.e. Hessian) of the loss function, with larger learning rates and smaller batch sizes yielding stronger regularization strengths \citep{keskar2016large, jastrzkebski2017three, jastrzebski2020break, cohen2021gradient}.
Meanwhile, theoretical works in simplified settings showed that this effect can be explained by Taylor-expanding the objective to third order, as such a calculation reveals that oscillating or fluctuating dynamics automatically induce curvature regularization \citep{blanc2020implicit,li2021happens,damian2021label,wen2022does,li2025adam}.
Building on this body of work, \citet{centralflows} recently showed that for several optimizers in the full-batch setting, the whole training trajectory on realistic neural nets is well-modeled by a curvature-penalized gradient flow, where the role of the hyperparameters is to modulate both the form and strength of the curvature penalty.
As a result, we now have a mathematical understanding of the learning rate in full-batch gradient descent, and are mostly free to instead study the simpler dynamics of gradient \textit{flow} plus a loss curvature penalty.\footnote{This perspective is reminiscent of the It\^{o} correction in stochastic calculus: after a nonlinear transformation, noise can contribute an additional deterministic drift. Likewise, stochastic or oscillatory optimization dynamics may be described by an effective flow on a modified loss.}
Other analyses have developed analogous characterizations for stochastic dynamics in more specialized settings \citep{pesme2021implicit, chen2024stochastic}.
Fully extending this characterization to stochastic and adaptive optimizers would give us a common language for reasoning about the implicit effects of optimization hyperparameters on the training trajectory.
It then remains to understand how these modifications to the training trajectory influence properties of the learned network (see \cref{openquestion:loss-curvature}).

\begin{figure}[ht]
  \centering
  \includegraphics[width=14cm]{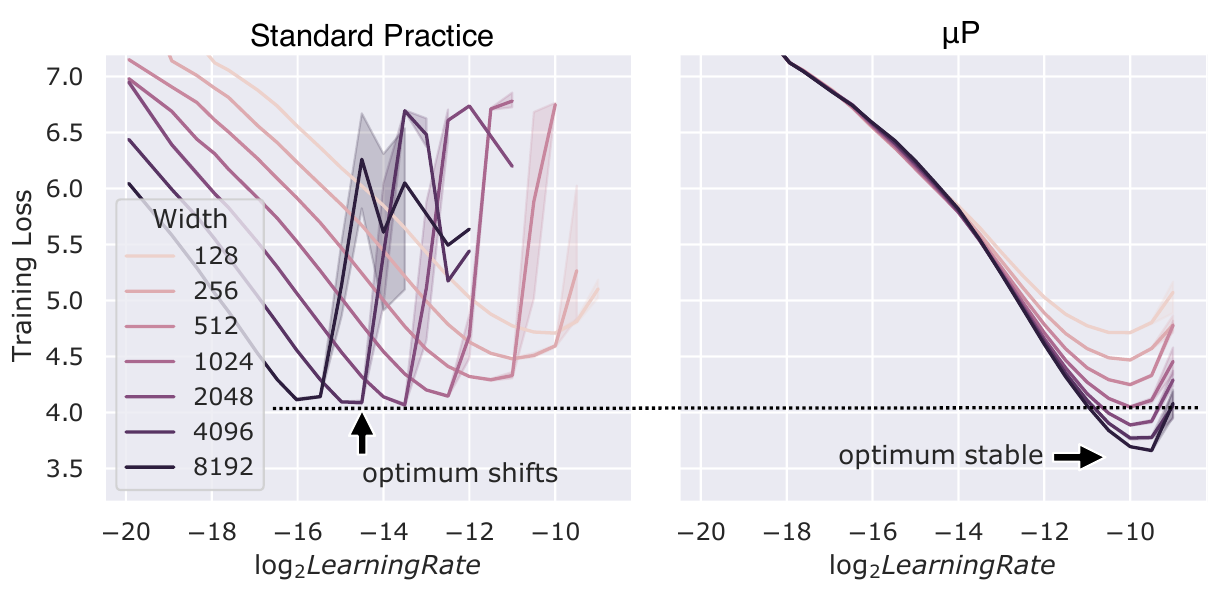}
  \caption{
  \textbf{The theory of network parameterization permits learning rate transfer across widths.}
  Transformers of varying widths trained on WikiText-2 under standard parameterization (left) and $\mu$P (right).
  Under standard parameterization, the optimal learning rate decreases as model width increases.
  Under $\mu$P, by contrast, the optimal learning rate remains nearly constant across widths, making it possible to predict the learning rate for wide networks from experiments on narrower, cheaper models.
  Adapted from \citet{yang2022tensor}.
  }
  \label{fig:hps-mup}
\end{figure}

\paragraph{Disentangling architecture hyperparameters from optimization hyperparameters.} 
There has been a highly successful line of work aimed at disentangling \textit{architecture} hyperparameters such as width, depth, and output multiplier (see the lazy/rich dichotomy in \cref{sec:reason-limits}), from \textit{optimization} hyperparameters such as the learning rate and initialization variance.
The Tensor Programs framework \citep{yang2021tensor, yang2023tensor} makes this separation explicit, writing hyperparameters such as the learning rate in the form $\eta = \eta_0 \cdot [\mathrm{width}]^{c}$, separating a scale-independent coefficient $\eta_0$ from a width dependent factor with exponent $c$.
% through the $abc$-parameterization, which writes a hyperparameter such as the learning rate as $\eta = \eta_0 n^{c}$, separating a scale-independent coefficient $\eta_0$ from the width dependence $n^c$ (see  \citep{yang2021tensor} for the full parameterization).
%
This line of work then asks: how can we set these exponents such that we retain interesting training behavior at infinite width?
A remarkable insight from this analysis is that all non-trivial and non-explosive scalings give one of two limiting behaviors, analogous to the rich/lazy dichotomy in \cref{sec:reason-limits}:
in the \textit{Neural Tangent Parameterization} (NTP), features are frozen during training, and in the \textit{Maximal Update Parameterization} ($\mu$P), features evolve.
Since feature learning is essential for most tasks, this analysis tells us that $\mu$P is the scaling to use, resolving how hyperparameters should scale with model width.
This understanding enables hyperparameter transfer: we can tune hyperparameters on small proxy models and then transfer them to large, production-size models, where they remain near-optimal when both models are sufficiently wide (\citep{yang2022tensor}; \cref{fig:hps-mup}).

At the same time, the theory underpinning this result is asymptotic and does not fully account for its empirical effectiveness.
In practice, models are trained at widths far smaller than the dataset size, and the usefulness of transfer depends on how \textit{quickly} optimal hyperparameters stabilize with width.
\citet{noci2024super} and \citet{ghosh2025understanding} and \citet{hayou2025proof} take steps toward closing this gap, providing evidence that a small set of spectral statistics stabilizes rapidly across widths under $\mu$P and approximately governs the optimal hyperparameters.
This scaling-centric approach to hyperparameters was later extended to depth scaling \citep{yang2023depth, bordelon2023depthwise, dey2025don}, and leveraging this approach with other scaling dimensions remains an important future direction (see \cref{openquestion:zero_hyperparameters}).

\subsection{Universal phenomena appear across settings and tasks}
\label{sec:reason-universal-behavior}

\begin{figure}[t]
    \centering
    \hfill
    \begin{subfigure}{0.5\linewidth}
        \centering
        \includegraphics[width=\linewidth]{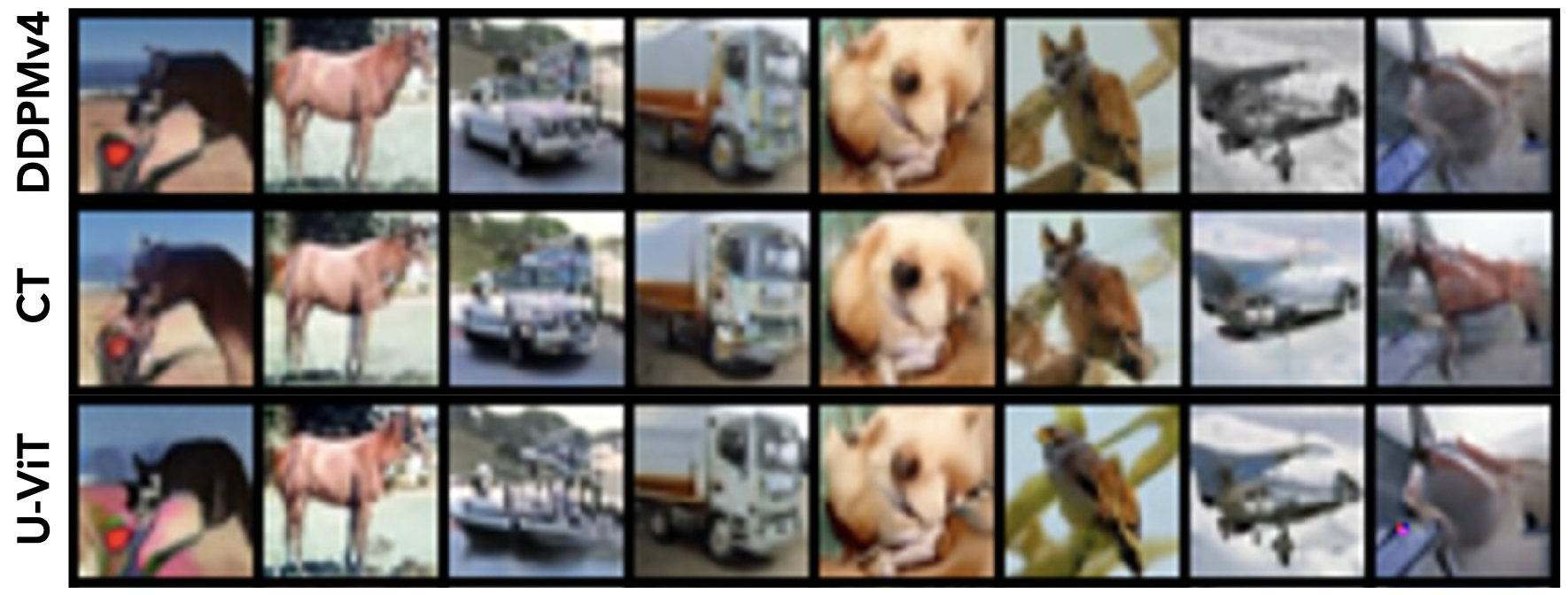}
        \caption{Universality across architectures}
    \end{subfigure}
    \hfill
    \begin{subfigure}{0.3\linewidth}
        \centering
        \includegraphics[width=\linewidth]{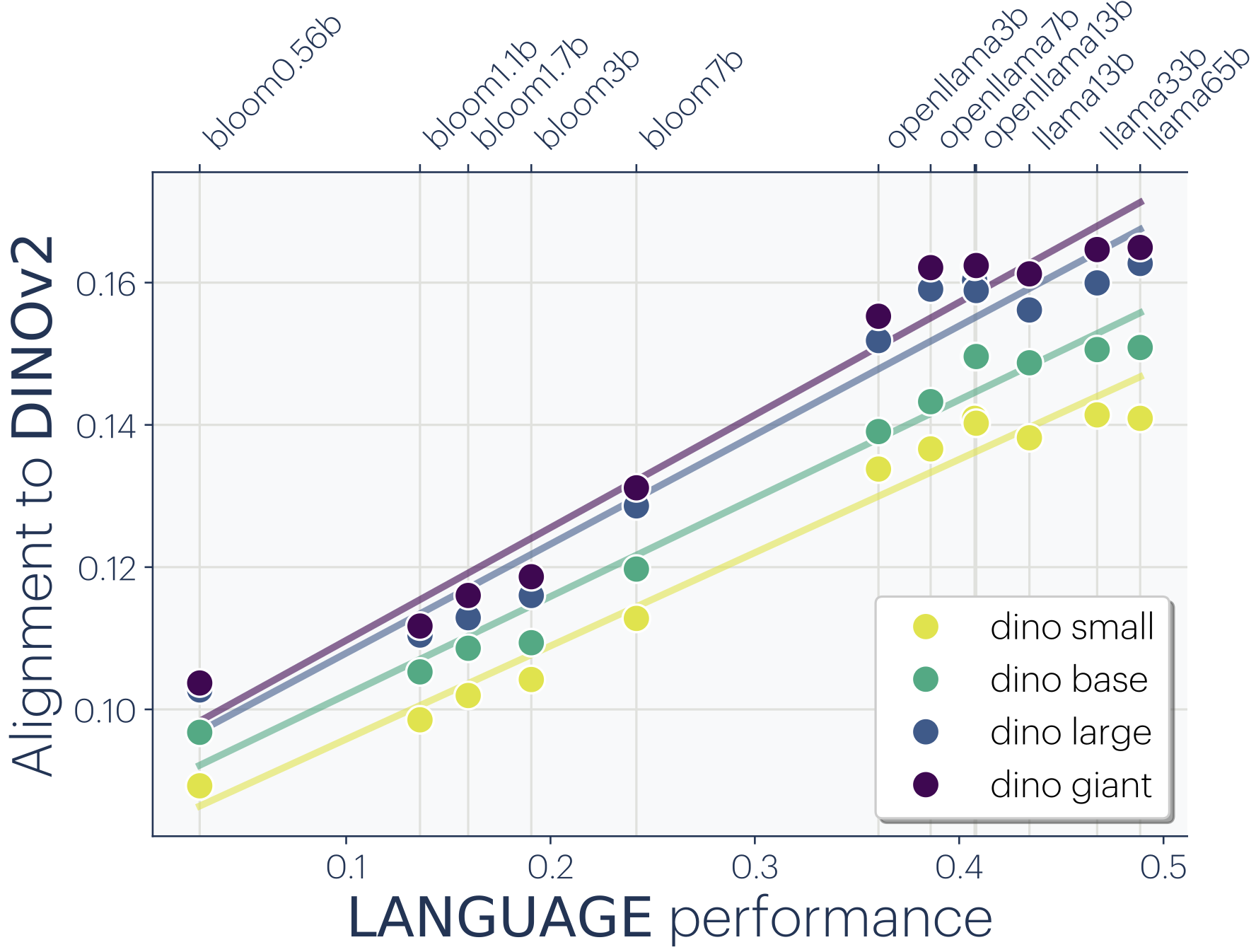}
        \caption{Universality across data modalities}
    \end{subfigure}
    \hfill\null
    \caption{\textbf{Universality across architectures and data modalities.} 
    (a):
    Different diffusion model architectures (from top to bottom: DDPM, a consistency model---both based on UNet---and U-ViT) converge to the same learned distribution and produce identical images when given the same input seed. Adapted from \citet{zhang2024emergence}.
    (b): As language model performance (horizontal axis) increases, their internal representations become increasingly similar to those of vision models, and more so for larger models (from yellow to purple lines). Adapted from \citet{huh2024platonic}.
    }
    \label{fig:cka_width_cvg}
\end{figure}

% \sout{Rather than a single recipe, deep learning is more akin to a cookbook \citep{balestriero2023cookbook} \jamie{not sure you need this citation}: there are many variations in architectures, training algorithms, and objectives.}

Deep learning is not a single recipe followed exactly every time: different systems use very different architectures, datasets, training algorithms, and objectives, with ingredients combined in creative ways.
This versatility has enabled successes on many tasks and modalities including vision, language, speech, time series, protein sequences, and games, but the resulting model diversity makes it less clear how to approach the development of scientific theory.
Do these diverse settings share deep commonalities we might hope to capture scientifically?

Here, we review a growing body of evidence that there are indeed \textit{universal phenomena} at play in these diverse settings.
This is good news for theory: when many different complex systems exhibit the same universal behavior, it suggests that a simple underlying explanation may exist.
% As an example, the central limit theorem explains how a wide variety of complex mechanisms give rise to Gaussian distributions, which only depend on the first two moments of the original source of randomness.
We highlight this universality through three different viewpoints: \textbf{(1)} different architectures
% \dhruva{and optimizers?}\florentin{Any good reference on that? Most optimization papers focus on the epsilon amounts of performance you save by changing the optimizer}
reach comparably good performance on many tasks; \textbf{(2)} different datasets share similar statistical properties; and \textbf{(3)} the learned representations and weights across different architectures and datasets are surprisingly alike.
This roughly echoes examples of universality in which disparate physical systems share deep commonalities %(such as having dynamics that minimize the action of a Lagrangian) \florentin{Commented out because this feels too general to capture what universality is}
or display similar behavior at large scales.%
\footnote{Universal behavior across physical systems can often be understood with the \textit{renormalization group,} a technique which formalizes the idea that, as one examines a system from a more and more zoomed-out perspective, most details ``wash out'' and only a handful of aggregate effects remain important. We note that another apt analogy for universality in deep learning, this one from biology, is \textit{convergent evolution}: species that ``solve similar problems'' tend to ``find similar solutions'' after many generations.}
We end by highlighting a few theoretical successes in modeling universal phenomena.

% Indeed, universality has been leveraged fruitfully in other fields to obtain theoretical tractability. A well-known example in physics is the study of phase transitions with the renormalization group: at large scales, microscopic details become irrelevant, with systems as different as boiling water and [{other example}] exhibiting nearly identical properties. Another well-known example is the central limit theorem, which explains how a wide variety of complex mechanisms give rise to Gaussian distributions, which only depend on the first two moments of the original source of randomness. This has important consequences for random matrices, where this so-called Gaussian universality has enabled several theoretical successes with quantitative predictions on natural data [cite].

% Universality may not be correlated with importance, and we must be careful not to attribute importance to certain attributes solely on account of the fact that they appear in many contexts. Thankfully, there is copious evidence that some of the essential features of deep learning have universal properties. 
% \florentin{I did not understand this point}

\paragraph{Universal inductive biases.}
% \jamie{oh, hm, I thought this was going to be (1) in the list above. is there a way you can make the structure here match that in the intro para, or make the intro para promise a breakdown like that you deliver in these paras?}
Performance on a given task is often robust to variations in architectures, training algorithms, and objectives, in the sense that many alternate choices still lead to models that can solve the task.
A well-known example is the choice between convolutional networks and transformers in computer vision tasks, which after much debate have been shown to obtain similar performance when matching compute, data size, and training recipes \citep{liu2022convnet,smith2023convnets}. 
%
% the input-output mappings learned by different models. 
% One instructive case study in this area comes from diffusion models. %Diffusion models can be sampled in a deterministic mode, and it has been noticed that in this mode, 
In diffusion models, this similarity has been further shown to hold at the level of input-output mappings, with transformers and UNets generating near-identical images when fed with the same noise samples \citep{zhang2024emergence}, as shown in \cref{fig:cka_width_cvg}.
These results strongly indicate that different architectures share similar inductive biases despite their apparent differences.
As a partial explanation, recent work has shown that assuming inductive biases towards locality and adaptivity to geometric structures leads to accurate quantitative predictions about the behavior of diffusion generative models \citep{kadkhodaie2024generalization, kambanalytic, niedobatowards}.

% \jamie{I wonder if you can make this para shorter to save space}

% These results suggest that different architectures and optimizers may share a large part of their inductive biases, and thus generalize for similar reasons.
% This greatly simplifies the search for a unified theory by freeing it from modeling the cornucopia of ingredients that enter the deep learning recipe.
% \jamie{nicely said}

\paragraph{Universal structure in data.}

% Training deep neural networks with stochastic gradient descent has emerged as a universal strategy to solve almost any task involving high-dimensional structured data. In addition to the observed convergence in inductive biases detailed above, practical design choices have also reduced in variation over the last few years, as transformers and Adam become defaults across problem settings.
% \jamie{do you need this para?}

The no-free-lunch theorem states that generalization on completely arbitrary data with a common learning strategy is not possible \citep{wolpert1996lack}.
Therefore, deep learning must rely on particular features of the data present across all datasets and modalities on which it succeeds \citep{goldblum2023no}.
For instance, many classes of images and audio signals share power-law spectral properties, sparsity patterns, and multiscale structures, and can be analyzed with general-purpose wavelet bases \citep{olshausen1996emergence,mallat1999wavelet}.
% \jamie{should we also cite https://www.cnbc.cmu.edu/~tai/readings/scaling/origin.pdf and https://www.nature.com/articles/381607a0 ?}].
A similar phenomenon in text data is the ubiquity of Zipf's law (word frequencies obey a power-law distribution) that holds over many natural and artificial languages \citep{li2002zipf,piantadosi2014zipf}. Hierarchical, compositional structure is also routinely used to model both images and text, which can sometimes be related through a common model \citep{cagnetta2024deep,sclocchi2025phase, cagnetta2025learning}. These shared statistical properties are a partial explanation for the ability of a single learning algorithm (say, a transformer trained with SGD) to tackle seemingly unrelated datasets, leaving only the finer-grained differences between them to be learned.

% \jamie{talk here about how it's fairly universal that nets learn ``simple'' fns first, then more complex? does seem like another avenue of theory worth mentioning at the end here in the ``recent work finding a partial explanation'' vein. for different models, ``simpler'' fns are always generally ``smoother'', even if the exact notion / eigenvals / etc. varies case by case.} \florentin{I agree with the intuition but it seems a bit too loose to be mentioned here---this would qualify as a universal phenomenon if one had a universal definition of simplicity}

% Evidence of universality has also been observed in the learned representations of data. It has been shown that networks trained to solve different tasks learn similar representations across training datasets (ImageNet and Places-365, \citet{lenc2015understanding}), objectives (supervised or self-supervised, \citep{bansal2021revisiting}), and modalities (vision or language, \citep{huh2024platonic}). The latter reference particularly highlights how this similarity increases with scale. As above, these analyses can be repeated at the level of weights \citep{guth2024universality}, showing that CNNs trained on non-overlapping class subsets of ImageNet learn similar features even at deep layers (while this similarity drops if one network is trained on random labels).

\paragraph{Universality in representations.} Going deeper in the internals of the network, it has been observed that representations learned by different networks can be similar across random initializations, widths, and architectures \citep{raghu2017svcca,kornblith2019similarity,bansal2021revisiting,huh2024platonic, moschella2022relative}. It has been shown that networks trained to solve different tasks learn similar representations across training datasets (ImageNet and Places-365, \citet{lenc2015understanding}), objectives (supervised or self-supervised, \citep{bansal2021revisiting}), and modalities (vision or language, \citep{huh2024platonic}). Furthermore, this similarity grows as model size and performance increase, hinting that neural activations converge towards a universal (``Platonic'') representation \citep{bansal2021revisiting, huh2024platonic}, as shown in \cref{fig:cka_width_cvg}.
In simplified settings such as random feature representations, this convergence is a consequence of the law of large numbers applied to the feature kernels \citep{rahimi2007random,guth2024rainbow}; in deep linear networks, it can be proven to arise from the implicit regularization of SGD~\citep{ziyin2025proof}; in more diverse settings, recent evidence suggests that the universality of representations may ultimately trace its origins to universal structure in data \citep{huh2024platonic,karkada2026symmetry}. Recent advances in {identifiability theory} \citep{hyvarinen2024identifiability} also have shown that representational convergence happens at the global optimum of unsupervised \citep{klindt2020towards}, self-supervised \citep{zimmermann2021contrastive} and supervised \citep{reizinger2024cross} objective functions under a suitable data generating process \citep{reizinger2025position}.
Several works have also shown empirically that this similarity can extend to the level of individual neurons \citep{li2015convergent,dravid2023rosetta,khosla2024privileged}. In some cases, similar representations have been found in both artificial neural networks and biological neural networks \citep{olshausen1996emergence, yamins2014performance, mcintosh2016deep}, though the extent of this correspondence remains controversial \citep{bowers2023deep}.
While a global trend towards similarity is emerging, it should be noted that the range of settings in which this convergence is observed, and its extent, are not fully known (see \cref{openquestion:comparing-models}). 
In particular, recent work has shown that this apparent convergence to universal representations depends crucially on the chosen comparison metric across similarities \citep{groger2026revisiting}. A growing literature is devoted to understanding which representation similarity metrics one should choose in different circumstances \citep{sucholutsky2023getting,klabunde2025similarity} and highlighting the cases where they can be unified \citep{harvey2024duality,williams2024equivalence}.
% It has been shown that for random feature representations, their increasing similarity is a consequence of the law of large numbers applied to their kernels \citep{rahimi2007random,guth2024rainbow}.

If the mechanisms learned by large models are indeed universal, this is very encouraging for theory: behavior shared across \emph{many} systems should depend primarily on the features common to \emph{all} such systems, and thus admit a description simpler than any particular model in isolation.
Moreover, if the internal structure of trained neural networks primarily reflects the structure of data, then in studying neural networks we may ultimately be studying the structure of data and its generating processes (see \cref{openquestion:theory_for_real_data}).
In particular, since language data comes directly from humans, understanding its structure may teach us something new and fundamental about ourselves.

\section{Relation to other perspectives}
% tag to be refactored
\label{sec:learning_mech_and_mechinterp}

There are several ongoing approaches to developing explanatory scientific theory of deep learning, each adopting a different perspective and using different sets of tools.
We believe that these perspectives are essentially all complementary: all either \textit{directly seek} a mechanics of learning or would \textit{symbiotically benefit} from one.

\paragraph{The statistical perspective.}
The rich tradition of classical learning theory remains influential today.\footnote{
The Simons Institute for the Theory of Computing has provided an important substrate for developing this statistical perspective of deep learning through \href{https://deepfoundations.ai/}{collaborations} and \href{https://simons.berkeley.edu/workshops/deep-learning-theory}{seminars}.
}
\citet{bartlett:2021-statistical-viewpoint} offer a lucid summary of its central framing: any statistical prediction method must balance \textit{expressivity} (to represent the richness of real data), \textit{complexity control} (to make the most of finite training data), and \textit{computational efficiency} (to yield practical algorithms).
% As \citet{bartlett:2021-statistical-viewpoint} beautifully put it, any statistical prediction method must balance three competing goals: it must be \textit{expressive} enough to represent the richness of real data, it must \textit{control} this complexity to make the most of finite training data, and it must be \textit{computationally efficient} to train.
It is apparent that deep learning is sufficiently expressive, but it is not clear how a good function is selected from this enormous function class, nor why simple gradient methods suffice to train such complex beasts.
The modern statistical viewpoint suggests two answers: deep learning has an \textit{implicit inductive bias} towards simple, well-generalizing functions \citep{teney2024neural,wilsonposition}, and despite their nonconvexity, the \textit{very high dimensionality} (overparameterization) of neural networks makes optimization easy.

These questions are important, and we believe these answers capture important principles.
% These questions are good ones, and we believe these answers are basically correct.
The challenge is now to specify how they are concretely implemented in the case of neural networks.
It is clear that doing so will require taking a close look at the nature of the training process.
Only once we have done so will we be able to back out \textit{how} this implicit bias arises and \textit{why} gradient methods suffice for optimization.
We do not believe these answers will be generic statements, but instead critically rely on important properties of deep learning and of natural data.
The statistical perspective thus leads naturally to a serious scientific study of the mechanics of training.

\paragraph{The information-theoretic perspective.}
A closely-related approach seeks to explain deep learning in terms of information-theoretic ideas.
In this view, learning is a process of extracting information from datasets, and a learning system works when it extracts information useful for prediction while discarding irrelevant information.
This perspective hopes to understand learning as \textit{compression} of the dataset into either the model's parameters or its hidden representations, with good generalization resulting when this compression is successful \citep{shwartz2017opening,xu2017information,blier2018description,lotfi2022pac,finzi2025compute}. The information-theoretic approach can be modified to account for computational aspects \citep{xu2020theory,finzi2026entropy}.

We find this perspective insightful, and it seems likely to us that a picture of this nature will hold.
As with the statistical perspective, a major remaining question is how to make this view concrete and actionable: how do the architecture and training process of deep learning interact to actually \textit{implement} this compression, and what factors make it more or less successful?
Doing this, too, will require taking a close look at the nature of the training process, the architecture, the data, and their interactions.
The information-theoretic perspective thus also leads naturally to a serious scientific study of the mechanics of training.

\paragraph{Physics of deep learning.}
This community descends from the older physics of machine learning lineage \citep{hopfield1982neural,amit1985spin,gardner1988space} and essentially seeks satisfying average-case theories of neural network learning \citep{lin2017does,zdeborova2020understanding,bahri2020statistical,michaudphysics,ringel2025applications}.\footnote{In the modern day, this community is mediated in part by recurring events at the \href{https://www.kitp.ucsb.edu/activities/deeplearning23}{Kavli Institute for Theoretical Physics}, the \href{https://aspenphys.org/event/theoretical-physics-for-artificial-intelligence}{Aspen Center for Theoretical Physics}, and the \href{https://leshouches2022.github.io/}{Les Houches School of Physics}, and organizations such as the \href{https://iaifi.org/}{NSF AI Institute for Artificial Intelligence and Fundamental Interactions} and the \href{https://www.physicsoflearning.org/}{Simons collaboration on the physics of learning and neural computation}.}
The close relationship between physics and machine learning was recognized by the \href{https://www.nobelprize.org/prizes/physics/2024/summary/}{2024 Nobel Prize in Physics}.
This approach is in line with (and has largely shaped) the perspective presented in this paper, and the project of this community is arguably the development of a mechanics of learning.
The challenge then is to clarify important problems and coordinate effort for efficient progress.

\paragraph{Perspectives from neuroscience.}
Several approaches to developing a science of the brain suggest approaches to developing a science of deep learning.
One approach starts from hypotheses about neural systems --- for example, that their computation amounts to some form of approximate probabilistic inference \citep{dayan1995helmholtz,friston2010free} or that they reduce sensory redundancy through \textit{efficient coding} \citep{barlow1961possible} --- and seeks to make deductions and predictions from these hypotheses.
Some of these predictions seem to hold suspiciously well in deep learning: see, for example, the case of edge-selective cells in the visual cortex \citep{olshausen1996emergence} and edge-selective receptive fields in convolutional networks \citep[e.g.][]{zeiler2014visualizing}.
Another approach termed \textit{systems neuroscience} seeks to directly decompose subsets of the brain into interpretable circuits and reverse-engineer the structure of their learned representations \citep{chung2021neural,bernardi2020geometry,kriegeskorte2008representational}.
This approach resembles mechanistic interpretability, which has adopted some of its methods and intuitions.
 
We expect and encourage this dialogue to continue, and it seems plausible that some of these high-level hypotheses about the brain --- e.g., that the brain admits at least a partial decomposition into interpretable circuits and that local circuits implicitly solve inference tasks --- will turn out to be true of deep learning.
The reasons these facts are true, if indeed they are, are surely bound up in the dynamical way learning actually happens.
A study of the mechanics of learning is thus important to the continued exploration of these ideas.
% We anticipate that a mature theory of the mechanics of deep learning would offer useful tools to theoretical and computational neuroscience.

\paragraph{Developmental interpretability/singular learning theory.}
This approach, which grew out of the mechanistic interpretability community, seeks first-principles predictive theories of neural network learning based on the singular learning theory framework of \citet{watanabe2009algebraic}, emphasizing a Bayesian perspective and aiming to understand training as a process of sequential phase transitions mediated by the geometry of the loss landscape \citep{hoogland2023developmental}.
We see this community as seeking the same goal we suggest here --- a fundamental mechanics of learning, and a rigorous foundation for interpretability --- but with a toolkit that differs from the other listed perspectives.
There is potential for fruitful cross-pollination and tool-sharing between these different approaches.

\paragraph{Science of deep learning.}
It has long been appreciated by practitioners that machine learning is largely a practice of trial and error and that it may be possible and beneficial to systematize it \citep{langley1988machine,gal2015science,rahimi2017nips_test_of_time,baraniuk2020science,biderman2026position}.
Indeed, much of the rapid empirical progress of the last decade resulted from systematic organization around agreed-upon benchmark tasks \citep{donoho2024data}.
Nonetheless, the training and application of large models remains more alchemy than science.
We believe that a fundamental mechanics of the learning process is the foundation on which this science will finally be built.

\subsection{\texorpdfstring{Learning mechanics $\rightleftarrows$ mechanistic interpretability}{Learning mechanics <-> mechanistic interpretability}}
\label{subsec:learning_mech_and_mechinterp}

We discuss mechanistic interpretability specially because there is a unique opportunity for cooperation.
% In this section, we discuss the symbiotic relationship we anticipate developing between learning mechanics and \emph{mechanistic interpretability}, another research program seeking low-level understanding of deep learning.
% 
Mechanistic interpretability aims to understand trained neural networks by identifying the internal mechanisms
--- features, circuits, and learned algorithms --- 
that give rise to their behavior. % \citep{sharkey2025open}.
% ^ we can totally keep that citation; just cut it bc we now cite that paper a few paras down in "calls for rigorous foundations"
%
At its core, this approach is guided by the belief that neural networks admit a  human-understandable, mechanistic description that can be uncovered through careful empirical reverse engineering.\footnote{The mechanistic interpretability community does not yet share a formal definition of what constitutes a ``mechanistic description,'' though see~\citet{geiger2025causal} for a recently proposed causal framing. Informally, many researchers proceed under a set of working assumptions: (1) that neural networks encode the state of internal computational variables in their activations, often referred to as ``features''; (2) that successive layers transform and combine these features in structured ``circuits''; and (3) that, taken together, these circuits implement algorithms that admit some level of human-understandable description.}
This approach has already borne fruit: many visually striking or interpretable mechanisms have been discovered in large models to date \citep{olah2020zoom,templeton2024scaling,engels2024not,gurnee2025when,lindsey2025biology}.%
\footnote{Mechanistic interpretability is deeply associated with \href{https://anthropic.com}{Anthropic} and the \href{https://www.alignmentforum.org/}{AI safety} and Effective Altruism communities, though it is increasingly pursued in academic labs. We also note that mechanistic interpretability has recently split into an \textit{ambitious} camp that hopes to develop full, explanatory scientific theory and a \textit{pragmatic} camp which is mostly interested in targeted interventions for particular cases. See also~\mbox{\cite{saphra2024mechanistic}} for a discussion of the origin of the term ``mechanistic interpretability'' and the dynamics of its community.}

This is a complementary perspective to our own and presents a wonderful opportunity for symbiosis.
At time of writing, mechanistic interpretability remains largely a qualitative science, more reliant on human-judged empirics than on compact mathematical principles or simple governing laws.
This is quite natural: semantically-meaningful functions resist mathematical characterization.%
\footnote{For example, try writing down a function that can classify dogs vs. cats from image pixel values. The difficulty of expressing such functions in mathematics is why we invented deep learning in the first place!}
On the other hand, a mechanics of learning would be quantitative by definition, but by the same token will be too low-level to answer important questions of semantic meaning on its own.
These approaches study the same system --- i.e., deep learning --- at different levels of abstraction, and so of course they can (and should) work together for mutual gain.
Calls for rigorous foundations for interpretability have been steadily growing \citep{sharkey2025open,joshi2026causality,greenspan2026towards}, and this is one thing learning mechanics can and should seek to help provide.
In turn, mechanistic interpretability offers learning mechanics a rich and growing tableau of empirical phenomena ripe for the development of explanatory mathematical theory.

\paragraph{Learning mechanics $\rightarrow$ mechanistic interpretability.}
We emphasize two complementary avenues through which learning mechanics can support mechanistic interpretability: formalizing core assumptions and explaining how mechanisms develop through training.

\emph{Formalizing core assumptions.} Learning mechanics can make explicit, formalize, and, where necessary, challenge the core and often implicit assumptions that guide interpretability research.
These include:
\begin{itemize}
  \item \emph{linear representability} --- that features correspond to meaningful directions in activation space \citep{mikolov2013linguistic, park2023linear, nanda2023emergent, marks2023geometry, jiang2024origins, csordas2024recurrent,garg2026many};
  \item \emph{locality} --- that features and circuits are localizable to particular subsets of model components \citep{meng2022locating, wang2022interpretability, conmy2023towards, arora2025language};
  \item \emph{sparsity} --- that individual features and circuits are activated or functionally relevant on only a small fraction of inputs ~\citep{cunningham2023sparse, bricken2023monosemanticity,hindupur2026projecting}; and
  % polysemanticity --- individual neurons may participate in multiple features \citep{olah2020zoom, cunningham2023sparse, bricken2023monosemanticity}; and
  \item \emph{compositionality} --- that complex network representations and computations arise from the composition of simpler, modular sub-mechanisms \citep{thorpe1989local,smolensky1990tensor,lepori2023break,schug2023discovering,ramesh2023compositional,okawa2023compositional}.
\end{itemize}
These core assumptions underpin the identification, isolation, and analysis of the internal mechanisms of trained neural networks in mechanistic interpretability research.
A mathematical theory of learning offers a way to clarify the regimes in which these assumptions hold, the conditions under which they fail, and the sense in which they can be derived from training dynamics and data statistics (see \cref{openquestion:what_are_features}). 
% 
% For example, recent theoretical work has begun to explain the formation of induction heads~\citep{reddy2023mechanistic}, how symmetries in pairwise co-occurrence statistics can induce linearly decodable representations \citep{karkada2026symmetry}, how neuron-specific feature representations can emerge from gradient-based training dynamics \citep{kunin2025alternating}, how polysemanticity can arise in toy settings \citep{elhage2022toy, hanni2024mathematical}, and how compositional structure in data can give rise to compositional circuits \citep{marchetti2026sequential}.

\emph{Explaining how mechanisms develop through training.} Mechanistic interpretability has generally prioritized describing \emph{what} mechanisms trained neural networks have learned, 
% while learning mechanics aims to explain \emph{how} and \emph{why} such structures develop during training. 
and there remains a rich opportunity for work which aims to explain \emph{how} and \emph{why} such mechanisms form in the first place.
There is already substantial interest within parts of the interpretability community in this dynamical/theoretical perspective, including work on the formation of induction heads~\citep{elhage2021mathematical, olsson2022context}, grokking and progress measures~\citep{nanda2023progress}, sudden phase transitions in circuit formation \citep{elhage2022toy, chen2023sudden, gopalani2024abrupt,park2024emergence}, and the research program of developmental interpretability~\citep{hoogland2023developmental,hoogland2025loss}, discussed earlier.
Our goal is not to replace these efforts but to encourage deeper engagement between mechanistic interpretability and the broader landscape of mathematically grounded ideas and tools in learning mechanics.
Echoing~\cite{saphra2022creationism}, we hope that
% In this sense, 
learning mechanics can play a role analogous to evolution in biology: just as \emph{``nothing in biology makes sense except in the light of evolution,''} the internal mechanisms of trained networks may be most naturally understood in the light of the processes that give rise to them.
%
% COMMENTED OUT FOR NOW -- @Eric if you have thoughts?
% \florentin{What are your thoughts on ``making simple, quantitative, statistical predictions/measurements''? As in, the way a physicist measures a system is different than the way a biologist would. Physicists would study average behaviors (what's the fraction of ``interpretable'' directions/circuits rather than highlighting the top few), how it scales with hyperparameters, and look for simple statements (it's logarithmic, linear, etc in width) rather than the sheer number of reported measurements or quantifying something for the sake of it (here's a table with the interpretability score of 37 architectures). Not an expert so tell me if it makes sense/is worth saying or not.}
% I like this point. 

\paragraph{Learning mechanics $\leftarrow$ mechanistic interpretability.}
Conversely, learning mechanics has been deeply influenced by the empirical discoveries of mechanistic interpretability, which often identify concrete phenomena that invite first-principles explanation.
Mechanistic interpretability places the structure of data at the center of its analyses, revealing settings in which the relationship between input structure and learned mechanisms is especially clear~\citep{nanda2023progress, shai2024transformers}.
By contrast, much of classical deep learning theory has relied on highly simplified data models, leaving a gap between theoretical predictions and behaviors observed in practice.
In this way, mechanistic interpretability helps bridge this gap by providing learning mechanics with concrete, well-defined targets for theoretical modeling.

Several such observations have already proven influential in stimulating work in learning mechanics, including the emergence of induction heads for in-context learning \citep{bietti2023birth, reddy2023mechanistic,nichani2024transformers}, the role of Fourier features in algebraic tasks \citep{morwani2023feature,kunin2025alternating,marchetti2026sequential}, and the geometry of features arising from the structure of correlations in the data \citep{engels2024not, prieto2025correlations, karkada2026symmetry}.
Just as the development of physics was often driven by empirical discoveries in adjacent fields, we expect progress in learning mechanics to be driven by theorists who take seriously empirical phenomena, including those uncovered by the mechanistic interpretability community, and seek to explain them.

% SECTION 4
\section{Reasons for skepticism and responses}
\label{sec:reasons_for_skepticism}

We have made a case that an ambitious mathematical theory of deep learning is possible and that developing this theory is a worthwhile endeavor.
This is far from a universal view, and so we now address common counterarguments that a theory of deep learning is either not possible or not a goal worthy of our effort.

\paragraph{Competent researchers have been trying to develop a theory of deep learning for decades, and we don't have one.
Surely if there was a theory, we would have already found it.}
It is true that machine learning theory is a field with a long history, and certain avenues for developing theory have been thoroughly explored.
Why should now be different?

There are several reasons for optimism.
First, the practical success of deep learning is comparatively recent, and we have a wealth of new empirical systems to study and mine for explainable phenomena.
Some of these phenomena, like the apparent convergence to universal representations discussed in \cref{sec:reason-universal-behavior}, were only revealed by the last few years of model scaling.
These developments have turned the search for a theory of deep learning from a mathematics into an empirical science (and one with no lack of interesting things to measure).
We now have much better means to ask questions and check our answers in a tight feedback loop.

Second, the field is much bigger: empirical successes have attracted researchers from physics, mathematics, neuroscience, and other adjacent fields, and so we have more and more diverse minds on the case. Third, it is worth noting that the development of major sciences has usually taken at least several decades, so we should not be too discouraged that we do not yet have all the answers.

% It is worth pointing out that the field \textit{is} currently making progress!
% A great deal of new ground has been broken in the last decade of deep learning theory.
% Whether due to the reasons above or some other cause, deep learning theory is not stuck and has clear avenues for future discovery.
% Of course, the question of whether this progress will continue and will eventually lead to a satisfactory mathematical understanding of deep learning is impossible to know, and our belief here is a matter of conviction as well as of evidence.
% This is the nature of any science, and only time will reveal the answer one way or another.

\paragraph{The objects currently understood from theory are very primitive compared to e.g. LLMs. Surely first-principles understanding of large models is too heavy a lift.}
Indeed, we expect that building up to LLMs will be a heavy lift and take considerable time.
The near-term hope is instead that some understanding of the basic building blocks of deep learning will prove useful even without a constructive theory that explains the whole model.
We can see this happening already in isolated pockets, including empirical scaling laws (\cref{sec:reason-measurable}), mathematical prescriptions for hyperparameter scaling (\cref{sec:reason-hps}), neural-tangent-kernel-based methods for data attribution \citep{park:2023-trak}, and theoretically motivated optimizers \citep{gupta2018shampoo,jordan2024muon,vyas2025soap}.
These ``local theories'' of small pieces of the deep learning stack are useful for hyperparameter scaling in large models, even though they are in no way comprehensive theories of the model.
% One might hope for similarly useful ``local theories'' that treat subjects like training instabilities, dataset selection and attribution, or the effect of normalization layers.

It is also important to stress that the identification of the right \textit{basic} objects in a field of science often makes it possible to ask \textit{applied} questions in a more sensible way.
Consider, for example, how the understanding that all matter is made of atoms underlies virtually all other basic science, and how knowledge of electromagnetism permits optical and radiological tools in countless applied disciplines.
As discussed in \cref{subsec:learning_mech_and_mechinterp}, we hope that learning mechanics can offer tools that adjacent fields such as mechanistic interpretability can apply to better carry out their work.
In this way, rigorous work on primitive objects can aid the applied science of large models even without a rigorous theory that builds all the way up.

% cut this bc we already talk about MI so much

% \paragraph{Deep learning is closer to biology than physics, and a trained neural network is more like a biological system than a physical one. A physics-inspired, first-principles scientific approach is thus misguided.}
% This is a common belief in the mechanistic interpretability community.
% It seems to us that deep learning is \textit{both:} a neural network trains and grows according to lawful basic principles, \textit{and} it grows into something complex like a living organism.
% Both perspectives are likely needed, and they should be mutually supportive.
% We give our vision of this relationship in \cref{subsec:learning_mech_and_mechinterp}.

\paragraph{What matters is a model's high-level behavior. Microscopic theories are too zoomed in to see this.}
Models' high-level behavior is indeed important.
How does this fit in with the lower-level sciences of deep learning?
We argue that deep learning may be studied at the level of physics, biology, or \textit{psychology,} with this last including the study of the model's capabilities, personality~\citep{betley2026training}, and goals.
It seems likely that study at all levels will be necessary.
Learning mechanics (the physics of deep learning) is the farthest from model psychology, with mechanistic interpretability (the biology) lying in the middle and connecting the two.%
\footnote{We note that these three levels of study of deep learning are roughly analogous to Marr's levels of analysis of a computational system: \textit{the physical implementation of the computation}, \textit{how the computation is performed algorithmically}, and \textit{what is being computed} \citep{marr2010vision}.}

\paragraph{We don't need a theory of deep learning, we need a theory of data.}
We think we need both: we need a theory of the structure in data and a theory of how a parameterized model learns it.
We touch on the necessity of developing a useful theory of data in \cref{sec:reason-universal-behavior,openquestion:theory_for_real_data}.
These are both part of the project of developing a mechanics of learning.

\paragraph{AI will understand itself before we do. Why try to build theory?}
This is a present concern for human intellectual endeavors across the board.
Our response here has three parts.
First, theory is \textit{already} useful, and will continue to be more impactful as it develops, so this scientific work is likely to make a near-term impact.
Second, it seems unlikely that AI working in isolation will suddenly and separately ``solve deep learning theory.''
It seems more likely that breakthrough progress in a transitory period will come from human scientists using or working with AI, and expert humans will remain in the loop.
Third, if one's goal is AI safety, some human oversight of AI systems will be necessary (unless one trusts the AIs to fully police themselves), and having a human-parseable theory of deep learning gives us a foot in the door.

% Lastly, and perhaps most contentiously, we as a group (with exceptions!) are generally more skeptical of the promise of superintelligence solving all current intellectual problems in the next decade than some other communities, and our work is partially a bet that AI will not understand itself before we at least make major progress on the problem.%
% \footnote{We note that there is a community divide over this: members of the mechanistic interpretability and AI safety communities tend to have shorter ``AI timelines'' than academic researchers of mathematical, statistical, or physical stripes.}
% \florentin{This is a difficult paragraph to write. We don't really give arguments, but it seems hard to do anyway, this is mostly a vibe check. But I do think it needs to be addressed, I just wish we could make a stronger case here.}

% SECTION 5
\section{Open directions in learning mechanics}
\label{sec:open_dirs}

It is important for any field, at any stage of development, to have a sense of its important open questions and goals.
In this section, we present a curated list of open directions which we expect can be solved by a theory of the mechanics of learning in the next decade.
These directions are loosely ordered by their connection to the lines of evidence introduced in \cref{sec:reasons_why}.
We hope this helps sharpen a shared research agenda.
For a longer catalog and a forum for community discussion, see \url{learningmechanics.pub/openquestions}.

\begin{opendirection}
[\emoji{crystal-ball}]
[What are simple, solvable models of genuinely deep, nonlinear learning?]
\label{openquestion:models_of_nonlin_learning}

As discussed in \cref{sec:reason-dynamics}, deep linear networks and kernel methods are the two main workhorse solvable models of learning mechanics.
The first captures nonlinear dynamics of the parameters, and the second learns nonlinear functions of the data.
While a few special cases of solvable models with both forms of nonlinearity are known, no unified framework has emerged.

Can we get the best of both worlds while maintaining some level of generality?
Is there a class of solvable model that captures both deep, nonlinear dynamics and nonlinear function learning?
Can such models illuminate new things about feature learning, the role of depth, optimization phenomena (e.g. progressive sharpening), and architectural innovations (e.g. normalization layers, residual streams, self-attention, and gated nonlinearities)?
Can they be usefully applied to modern learning paradigms like self-supervised learning, reinforcement learning, and denoising diffusion?
\end{opendirection}

\begin{opendirection}
[\emoji{elephant}]
[What would a theory capable of capturing natural data look like?]
\label{openquestion:theory_for_real_data}

Deep neural networks find and exploit structure in natural data.
This means that the structure of the data must somehow enter into our theories.
What is this structure, and how do we find it?

Despite the complexity of data, in many cases models appear to derive their learning signal from a small set of sufficient statistics.
What are these minimal data statistics, and how do they enter into a predictive theory of what the model learns?
Are these statistics different for different models and at different stages of training?
Can we describe the relevant structure in a dataset in terms of a model with free parameters found via an empirical fit?
\end{opendirection}

\begin{opendirection}
[\emoji{abacus}]
[Does deep learning implicitly minimize some notion of functional complexity?]
\label{openquestion:functional-complexity}

Deep networks trained by conventional optimizers are widely believed to have some sort of bias towards learning simple functions.
This idea has surfaced many times under different names (e.g. implicit regularization, maximum margin bias, simplicity bias, and spectral bias), but has only been characterized precisely in highly specific settings, and a general picture has not been found.    
Do deep neural networks broadly seek to minimize some precise notion of complexity among functions with low loss?
    
If so, what is the appropriate notion of complexity --- Kolmogorov, circuit, weight norm, or something else?
In what settings or limits is this minimization exact, and when is it only approximate?
Do the sparse features and circuitry studied by mechanistic interpretability naturally emerge as the solution to this minimization problem?
\end{opendirection}

\begin{opendirection}
[\emoji{microscope}]
[How do we formally define the \textit{features} learned by neural networks?]
\label{openquestion:what_are_features}

Mechanistic interpretability seeks to identify and disentangle the \textit{features, circuits, and mechanisms} learned by neural networks.
Can these concepts be given precise mathematical definitions grounded in first principles?
What formal structures naturally emerge from such a definition?
Can we use these notions to evaluate and formalize central assumptions of mechanistic interpretability, including \textit{linear representability, locality, sparsity, }and\textit{ compositionality,} as discussed in \cref{sec:learning_mech_and_mechinterp}?
How do these ideas connect with the less semantically-meaningful --- but more precise --- rich vs. lazy picture of feature learning discussed in \cref{sec:reason-limits}?
\end{opendirection}

\begin{opendirection}
[\emoji{infinity}]
[Are finite neural networks properly understood as approximations to infinite limits?]
\label{openquestion:discretization}

In \cref{sec:reason-limits}, we articulated the \textit{Discretization Hypothesis,} which states that finite neural networks are simply discretized approximations to infinite networks, analogous to how a spatiotemporal discretization is used to numerically approximate the solution to a differential equation. For network width, the limiting continuous object is the measure of neuron activity in hidden layers, while finite depth in a residual network can be viewed as a discretization of a neural SDE or ODE.
Small step sizes can render stochastic optimization algorithms approximately equivalent to some kind of flow. In this view, increasing model size (and decreasing learning rate while commensurately increasing step count) serves essentially to improve model performance by decreasing discretization error, at the price of additional computation.
Is this the right way to understand width, depth, learning rate, and other finite hyperparameters in deep learning?
What does the limiting continuum system look like?

\end{opendirection}

\begin{opendirection}
[\emoji{broom}]
[Can we understand and eliminate all hyperparameters?]
\label{openquestion:zero_hyperparameters}

In \cref{sec:reason-limits,sec:reason-hps}, we outlined a research program in which hyperparameters are systematically analyzed, disentangled, and in some cases removed by taking appropriate limits.
How far can this program go?
Can we reach zero hyperparameters, or are some hyperparameters irreducible?
If we eliminate all hyperparameters, what remains?%
% Is there a model of learn
\end{opendirection}

\begin{opendirection}
[\emoji{triangular-ruler}]
[Can we predict scaling law exponents \textit{a priori}?]
\label{openquestion:predict_powerlaw_exponents}

As discussed in \cref{sec:reason-measurable}, large models exhibit robust power-law scaling of loss with respect to model size, data, and compute. 
The observed exponents are nontrivial: they do not appear to be simple fractions which might result from elementary dimensionality arguments.
It is widely believed that these values are driven largely by structure latent in the dataset, but may also depend on details of the architecture and optimizer.
While many explanations for scaling laws have been proposed, a decisive test of any such theory is its ability to predict these exponents quantitatively from first principles.
At present, no framework can robustly do so across realistic settings.
Can we develop a theory of scaling laws that both explains why power laws arise and predicts their exponents \textit{a priori}?
What measurements of the dataset, architecture, and optimization are required to do so?
\end{opendirection}

\begin{opendirection}
[\emoji{roller-coaster}]
[How does loss curvature interplay with architecture, features, and generalization?]
\label{openquestion:loss-curvature}
As discussed in \cref{sec:reason-measurable,sec:reason-hps}, a significant feature of deep learning optimization is that the optimizer implicitly regularizes the curvature (i.e. Hessian) along its trajectory, by steering towards regions of the loss landscape with lower curvature.  While progress has been made on formalizing this effect using curvature-penalized gradient flows, it remains unclear how these curvature dynamics relate to other concerns in deep learning theory.  Why does the curvature tend to rise in the absence of any such implicit regularization, and can this ``progressive sharpening'' be attributed to certain properties of the architecture or data distribution?  How does the implicit curvature regularization affect the features that are learned?  Why does it sometimes lead to improved generalization?
\end{opendirection}

\begin{opendirection}
[\emoji{racing-car}]
[What makes for a good optimizer in deep learning?]
\label{openquestion:good-optimizer}

It remains fundamentally unclear why some deep learning optimizers work better than others.
Why do adaptive methods, such as Adam and Muon, consistently outperform simpler alternatives like SGD when training large language models?
How does adaptive preconditioning in these optimizers interact with a network's architecture and loss landscape to lead to faster, more stable training?
Can we identify fundamental principles that explain the success of modern optimizers, predict when they will fail, and guide the design of new ones?

\end{opendirection}

\begin{opendirection}
[\emoji{dancing-women}]
[In what sense do large models trained differently learn similar representations?]
\label{openquestion:comparing-models}

In \cref{sec:reason-universal-behavior}, we discussed evidence that large models trained from different random seeds --- and sometimes even with different widths, architectures, data, or objectives --- tend to learn similar internal representations.
A precise version of this claim would be very powerful: understanding how representation learning is \emph{universal} would give us confidence that theory developed for one model and setting transfers to many others.

The central difficulty here is methodological: how do we assess ``similarity''?
There is no single way to compare high-dimensional representations --- metrics based on kernel alignment, nearest-neighbors, model stitching, and more compare different aspects of representation geometry. Which ones are stable across training regimes? What is the appropriate metric that quantifies this similarity? What is the largest range of experimental settings under which convergence is observed --- what are the representation universality classes?
% can yield different results.
% 
% When do such notions agree, and which aspects of representation geometry are stable across training regimes?
% 
% Can we develop principled similarity metrics that permit sharp, falsifiable statements about representational similarity?

\end{opendirection}

\section{How to get involved in the development of learning mechanics}
\label{sec:how_to_get_involved}

It is always difficult to start doing research in a new field.
Consequently, we would like to make it as easy as possible for newcomers to get started.
In this section, we extend a hand with some encouragement and advice.

There is no specific academic background required to do useful work in this field.
Well-regarded researchers in deep learning theory come from backgrounds in physics, mathematics, computer science, neuroscience, statistics, and more.
Moreover, knowing another field well is useful, and established ideas from other fields can be applied to deep learning in some form or another, as the diversity of perspectives on deep learning theory attests (\cref{sec:learning_mech_and_mechinterp}).
Good things grow from cross-pollination.
A firm grasp of undergraduate mathematics, a familiarity with deep learning, and a desire to learn are the only definite prerequisites.

\subsection{Tenets for getting started}
If you want to join this field, you are more than welcome.
While there is no single correct way to craft theory, there are plenty of pitfalls that many of us encountered when starting out.
To help avoid some of them, we have compiled a shortlist of guiding principles for doing research in this field.
These tenets are not intended to maximize your number of citations in the short term, and following them may involve some swimming against the current of academia.
Instead, they are intended to maximize your impact in the long term and your ability to integrate and contribute to the community.

\begin{enumerate}

    \item
    \textbf{Do experiments frequently.} 
    As discussed in \cref{sec:reason-measurable}, deep learning is a field where the cost of doing experiments is relatively low, with a fast turnaround time.
    Use that to your advantage!
    Experiments serve to check assumptions, inform theoretical models, reveal the limitations of a theory, peer beyond the cases that a theory covers, and surface interesting phenomena to study in the first place.
    Try to include experiments in every paper, and make them as simple and revealing as possible.
    
    \item
    \textbf{Simplicity and insight matter more than technical complexity.}
    If you want to do work that is useful for others, they need to understand it, not merely be impressed by it.
    Take the time to simplify your findings, identify the underlying intuition, and check with simple experiments.
    A useful idea for thinking about a \textit{type} of problem is generally more valuable than a difficult theorem or a solution to any \textit{particular} problem, so emphasize these useful ideas when you present your work.
    This will make your results more accessible and easier for others to extend and apply.
    Your conference reviewers may disagree with this philosophy --- the conference system tends to reward technical complexity and undervalue simplicity --- but it will make your work more impactful in the long run.
    
    % \item
    % \textbf{Share failures and half-baked work.}
    % Science is slowed down when scientists only communicate completed, successful ideas.
    % Especially when starting, you should share your failures and half-baked ideas, ideally on a public platform like a blog.
    % This will help you get feedback, and writing up unfinished projects will make it easier to revisit them later. \florentin{I feel uneasy about the title+1st sentence---don't we feel like ML conferences are bloated and have mainly low-quality work? It seems important to think about who to share an idea with and where based on how refined/of general interest it is in order to de-saturate our communication channels, so I would emphasize the blog aspect a bit more}

    \item
    \textbf{Value scientific understanding over state-of-the-art performance.}
    Applied deep learning is a field of engineering whose progress is measured by benchmarks.
    For scientists of deep learning, the game is different: your contribution is gauged by your contribution to collective understanding.
    It is easy to feel pressure to tack on some engineering benefit in a scientific paper to make the paper seem more relevant and timely, but doing so usually dilutes the paper's scientific contribution without really affecting practice.
    Of course, fundamental science should eventually improve state-of-the-art performance, and when this naturally falls out of the science, it is a powerful way to demonstrate what has been understood.%
    \footnote{For a good example of this, see \citet{yang2022tensor}.}
    When it does not, though, there is no need to force it: set benchmarks aside for the time being and seek understanding on its own terms.

    \item
    \textbf{Don't try to do it alone.}
    Deep learning is a field with a lot of history and many known results, and guidance from a live human will help you.
    You can get pretty far from reading material online and talking to AI, but it doesn't replace human collaboration and mentorship.
    You should seek out other people interested in this area, ask for their feedback, and ask to work with them.
    A weak corollary to this is: when one exists, watch the talk version of a paper in addition to reading it.
    A great deal more of the nuance of a project is conveyed in a live presentation.
    The \href{http://physicsmeetsml.org/}{Physics Meets ML} and \href{https://www.physicsoflearning.org/webinar-series}{Physics of Learning} series are good places to look for talks on what we here call learning mechanics.

    \item
    \textbf{Try a few different problems before going deep into one.}
    Deep learning theory has so many open questions that you probably won't know where to start.
    That's okay --- just jump in, and feel free to change problems a few times early on.%
    \footnote{Most authors of this paper did this, as you can see from our respective research records!}
    Knowing multiple areas is essential to having high-level ideas anyways, and working in an area is the best way to learn it.
    
    \item
    \textbf{Invest in fundamental tools and techniques.}
    Compared to more established fields of science, mathematics, and engineering, deep learning is very young.
    These other fields have identified deep ideas and developed powerful tools applicable to broad classes of problem.
    Learning these fundamental tools comes in handy when similar problems arise in deep learning.
    For example, statistical physics and random matrix theory have powerful tools for thinking about high-dimensional interacting systems, and these tools are of great use in learning mechanics when one takes infinite limits.
    Many of the central ideas of classical optimization theory make regular appearances in the study of neural network optimization.
    Other concepts from statistical signal processing, such as wavelet decompositions, graphical models, and information theory, have been comparatively less leveraged so far, but may prove to be useful and complementary tools of learning mechanics.

\end{enumerate}

We will put as much useful introductory material as we can on \websiteurl, and we encourage discussion in the comments there.
We also encourage taking a crack at the open directions in \cref{sec:open_dirs}.
Work hard, have fun, and best of luck --- we hope to see a great deal more fundamental science of deep learning in the next few years!

\begin{flushright}
    \small
    \textit{``Wir m\"ussen wissen. Wir werden wissen.''}\\
    ``We must know. We will know.''\\
    --- David Hilbert, 1930
\end{flushright}

\newpage
\section*{Acknowledgements}
We are grateful for feedback on this paper from many people from several overlapping groups: researchers working on the theory of machine learning, researchers working on AI safety and mechanistic interpretability, practitioners and applied deep learning scientists, neuroscientists, and physicists.
This includes
Alberto Bietti,
Alex Infanger,
Alex Williams,
Amil Dravid,
Andrew Gordon Wilson,
Anthony Thomas,
Avrajit Ghosh,
Bin Yu,
Bruno Loureiro,
Chandan Singh,
Clémentine Dominé,
Dan Yamins,
David Berman,
David Klindt,
Denny Wu,
Ev Gunter,
Honam Wong,
Itay Lavie,
Jacob Yates,
Jacob Zavatone-Veth,
Jeff Gore,
Jesse Hoogland,
Jiechao Feng,
Jingfeng Wu,
Kaden Tro,
Kaiyue Wen,
Lauren Greenspan,
Lawrence Chen,
Lenka Zdeborová,
Lily Stelling,
Lukas Bongartz,
Marco Gherardi,
Nina Miolane,
Noa Rubin,
Peter Bartlett,
Raymond Fan,
Samyak Jain,
Soufiane Hayou,
Sultan Daniels,
Wanyu Lei,
Weijie Su,
Yasaman Bahri,
Yuandong Tian,
and
Zohar Ringel.
We are grateful to Imbue for funding the retreat at which this project originated.

% \newpage
\bibliography{refs}

\end{document}